\documentclass[10pt,twocolumn,letterpaper]{article}

\usepackage[pagenumbers]{cvpr} 

\usepackage[utf8]{inputenc} 
\usepackage[T1]{fontenc}    
\usepackage{url}            
\usepackage{booktabs}       
\usepackage{nicefrac}       
\usepackage{microtype}      
\usepackage{xcolor}         
\usepackage{amsmath,amsfonts}
\usepackage{algorithmic}
\usepackage{algorithm}
\usepackage{array}
\usepackage{textcomp}
\usepackage{stfloats}
\usepackage{verbatim}
\usepackage{graphicx}
\usepackage[english]{babel}
\usepackage{tabularray}
\usepackage{multirow}
\definecolor{mygray}{rgb}{0.9,0.95,1}
\usepackage{booktabs}
\usepackage{tikz}
\usetikzlibrary{tikzmark,calc}  

\definecolor{cvprblue}{rgb}{0.21,0.49,0.74}
\usepackage[pagebackref,breaklinks,colorlinks,allcolors=cvprblue]{hyperref}

\def\paperID{} 
\def\confName{CVPR}
\def\confYear{2026}

\title{Decoupling Bias, Aligning Distributions: Synergistic Fairness Optimization for Deepfake Detection}

\author{
Feng Ding$^{1}$,
Wenhui Yi$^{1}$,
Yunpeng Zhou$^{1}$,
Xinan He$^{1,2}$,
Hong Rao$^{1}$\thanks{Corresponding author.},
Shu Hu$^{3}$\footnotemark[1] \\[2pt]
$^{1}$Nanchang University {\tt\small \{fengding, raohong\}@ncu.edu.cn,}\\
{\tt\small \{408000240010, yunpengzhou, shahur\}@email.ncu.edu.cn}\\
$^{2}$Guangdong Provincial Key Laboratory of Intelligent Information Processing,\\
\phantom{$^{2}$}Shenzhen Key Laboratory of Media Security,\\
\phantom{$^{2}$}and SZU\textendash AFS Joint Innovation Center for AI Technology, Shenzhen University\\
$^{3}$Purdue University\ {\tt\small hu968@purdue.edu}
}

\begin{document}
\maketitle
\begingroup
\renewcommand\thefootnote{}\footnotetext{This paper has been accepted by CVPR 2026.}
\endgroup
\addtocounter{footnote}{-1}
\begin{abstract}
Fairness is a core element in the trustworthy deployment of deepfake detection models, especially in the field of digital identity security. Biases in detection models toward different demographic groups, such as gender and race, may lead to systemic misjudgments, exacerbating the digital divide and social inequities. However, current fairness-enhanced detectors often improve fairness at the cost of detection accuracy. To address this challenge, we propose a dual-mechanism collaborative optimization framework. Our proposed method innovatively integrates structural fairness decoupling and global distribution alignment: decoupling channels sensitive to demographic groups at the model architectural level, and subsequently reducing the distance between the overall sample distribution and the distributions corresponding to each demographic group at the feature level. Experimental results demonstrate that, compared with other methods, our framework improves both inter-group and intra-group fairness while maintaining overall detection accuracy across domains. The code is available at https://github.com/ywh1093/Fairness-Optimization.
\end{abstract}    
\section{Introduction}
\vspace{-2mm}

Deepfake, a combination of ``deep learning'' and ``fake'', refers to the use of deep learning techniques to replace faces in real images or videos with others, thereby generating fabricated identities. Due to their high fidelity, synthetic media can evade human perception and are maliciously exploited in financial fraud, judicial evidence manipulation, and misinformation dissemination~\cite{tolosana2020deepfakes,ding2021anti, zhou2025breaking,ding2026diffface}. The convergence of technological sophistication and societal impact has made Deepfakes a critical security threat requiring urgent governance in the AI era~\cite{ding2022securing,tolosana2020deepfakes,citron2019deepfakes}.

\begin{figure}[t]
\centering
\includegraphics[width=0.9\linewidth]{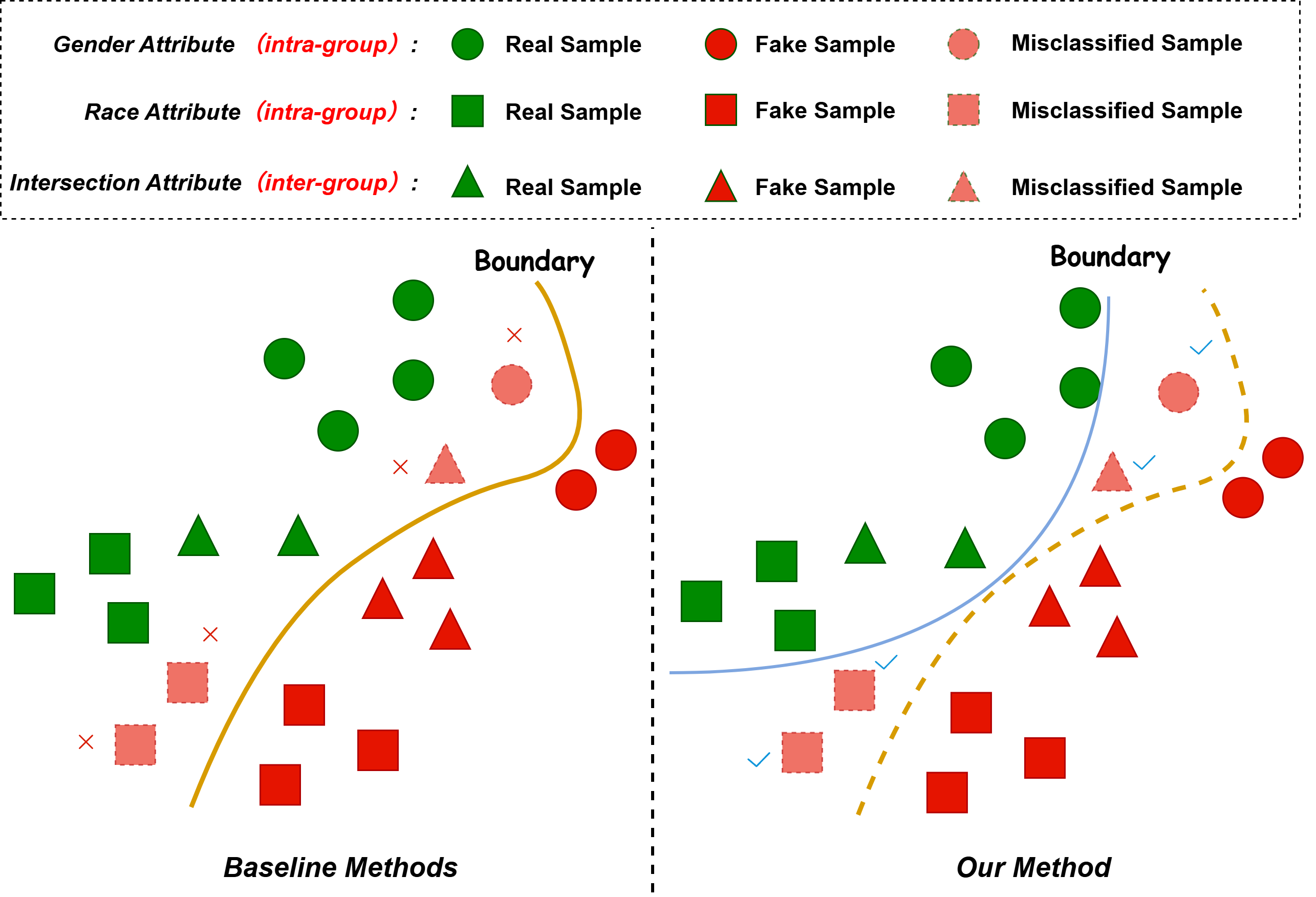}
\vspace{-1mm}
\caption{Example illustrating our proposed fairness enhancement strategy. The baseline method may lead to unfair model performance across intra-group and inter-group.
Our proposed method effectively addresses this issue by more balanced handling of different sensitive attributes.}
\label{fig:intro}
\vspace{-3mm}
\end{figure}

To mitigate the risks posed by Deepfakes, numerous detection methods have been proposed in recent years~\cite{masood2023deepfakes,leibowicz2021deepfake,passos2024review,ye2024decoupling,fan2025generating, zhou2026simplicity}. Nevertheless, fairness has often been overlooked, resulting in inconsistent performance across demographic groups defined by gender, race, and age~\cite{krishnan2020understanding,albiero2021gendered}. A principal driver of these disparities is dataset imbalance: large benchmarks frequently mirror societal or cultural skews, with Caucasian faces and certain genders overrepresented (\textit{e.g.}, FF++ dataset~\cite{rossler2019faceforensics++}). Under standard empirical risk minimization, models are biased toward the majority distribution, producing unequal error rates across groups. For example, detectors commonly achieve higher accuracy on lighter than on darker skin tones~\cite{trinh2021examination}.

Early efforts~\cite{nadimpalli2022gbdf, wang2022fairness} addressed fairness from the dataset side by balancing demographic distributions through resampling. Subsequent work~\cite{ju2024improving} shifted to algorithm-level objective design and loss reweighting to reduce bias, yet generalization remained insufficient. Disentanglement was then explored to separate demographic features from domain-agnostic forgery cues, which improved fairness generalization but reduced detection accuracy~\cite{lin2024preserving}. Overall, existing strategies either fail to generalize or degrade detection accuracy, which motivates our approach that improves both inter- and intra-group fairness while maintaining detection performance.

As shown in Fig.~\ref{fig:intro}, performance disparities across demographic groups are effectively mitigated by the proposed method, while overall detection accuracy is maintained. Specifically, this paper proposes an innovative synergistic fairness optimization framework that integrates structural fairness decoupling and global distribution alignment. In the structural fairness decoupling stage, we compute fairness index to identify channels highly correlated with demographic groups and decouple a certain percentage of channels, thereby reducing the model’s reliance on these sensitive attributes that may introduce bias into the model. In the global distribution alignment stage, we enhance model fairness by minimizing the distance between the overall feature distribution and the distributions corresponding to each demographic group. The purpose of the second stage is to derive common sense from the decoupled features, thereby reducing the discrepancies across different domains.

In sum, the contributions of our work are three fold:
\begin{itemize}
    \item This paper introduces a structural fairness decoupling module to dynamically identify and decouple channels that are highly correlated with sensitive attributes. By reducing the model’s reliance on sensitive attributes, the module narrows performance disparities across groups and improves both inter-group and intra-group fairness.
    \item This paper proposes a global distribution alignment module that aligns the distributions between the overall sample and  each demographic group. With the decoupled demographic-insensitive features, this module can extract common sense from them to reduce discrepancies across different domains, thereby further enhancing fairness generalization.
    \item This method outperforms state-of-the-art approaches in improving both inter-group and intra-group fairness while maintaining overall detection accuracy during  deepfake detection, as demonstrated in extensive experiments on various leading deepfake datasets.
\end{itemize}

\section{Related Work}
\vspace{-1mm}
\subsection{Deepfake Detection}
\vspace{-1mm}
Based on prior research, deepfake detection methods can be broadly categorized into four main directions: biological feature–based detection, signal-level artifact analysis, data-driven deep neural network (DNN) classification, and forgery-trace–oriented approaches. Early studies~\cite{lyu2022deepfake} focused on extracting biological or physiological cues from facial regions to expose tampering traces. Subsequently, researchers~\cite{wang2022deepfake} found that signal-level inconsistencies such as frequency or compression artifacts were more reliable for identifying synthetic content. Current deepfake detection research is predominantly grounded in convolutional neural network (CNN)–based classification frameworks~\cite{ramachandran2021experimental,guo2022robust,pu2022learning,ding2024disrupting,ye2024decoupling}, which learn discriminative forgery features through end-to-end training. As forgery techniques have evolved in complexity, recent studies have advanced toward forgery-trace–oriented approaches. For instance, Cao et al.~\cite{cao2022end} introduced a forgery detection framework that relies on reconfigurable classification learning. There are also detectors developed with large language models \cite{he2025vlforgery}.

\subsection{Stage-Wise Fairness Optimization Methods}
\vspace{-1mm}
Previous studies have documented significant demographic biases in deepfake detectors across gender, age, and race~\cite{masood2023deepfakes, pu2022fairness, xu2024analyzing, trinh2021examination, hazirbas2021towards}. To mitigate these biases, prior work can be organized into pre-processing, in-processing, and post-processing stages~\cite{ding2025fairadapter, lin2024preserving, cheng2025fair}.

\noindent
\textbf{Pre-processing methods.}~\cite{nadimpalli2022gbdf, wang2022fairness} rebalance data via resampling or cross-group synthesis to alleviate distribution skew. These methods show limited adaptability to evolving generative attacks and sparsely observed intersectional combinations.

\noindent
\textbf{In-processing approaches.}~\cite{lin2024preserving, ju2024improving, little2022fairness, roh2020fr, sarhan2020fairness, he2024redundant, ding2025fairadapter} incorporate adversarial debiasing, risk-sensitive objectives, or disentanglement modules to reduce the dependence of predictions on sensitive attributes. These mechanisms can inadvertently suppress discriminative forgery cues and degrade accuracy or robustness.

\noindent
\textbf{Post-processing methods.}~\cite{tifrea2023frappe, hardt2016equality, pleiss2017fairness} calibrate thresholds or align output distributions conditioned on sensitive attributes. Their effectiveness is constrained by residual representation bias and instability under cross-domain shifts.

\section{Method}
\vspace{-1mm}
\subsection{Overview}
\vspace{-1mm}
To address fairness concerns in deepfake detection, we propose a novel method that improves both intra-group and inter-group fairness without sacrificing detection performance and demonstrates a notable ability to generalize fairness in this section.

\noindent
\textbf{Problem Definition.} Given a training dataset \( D_{\text{sensitive}} = \{(x_i, y_i, a_i)\}_{i=1}^m \) with size m, where \(x_i\) represents an input image, \(y_i\in\{0 : real, 1 : fake\}\) indicates identity label, and \(a_i\) represents a sensitive attribute (single or intersectional). Our goal is to train a fair deepfake detection model \(f_{\theta}\) using \(D_{\text{sensitive}}\) that both accurately detects Deepfakes and maintains fairness across attribute groups. 

\begin{figure*}[ht]
  \centering
  \includegraphics[width=\textwidth]{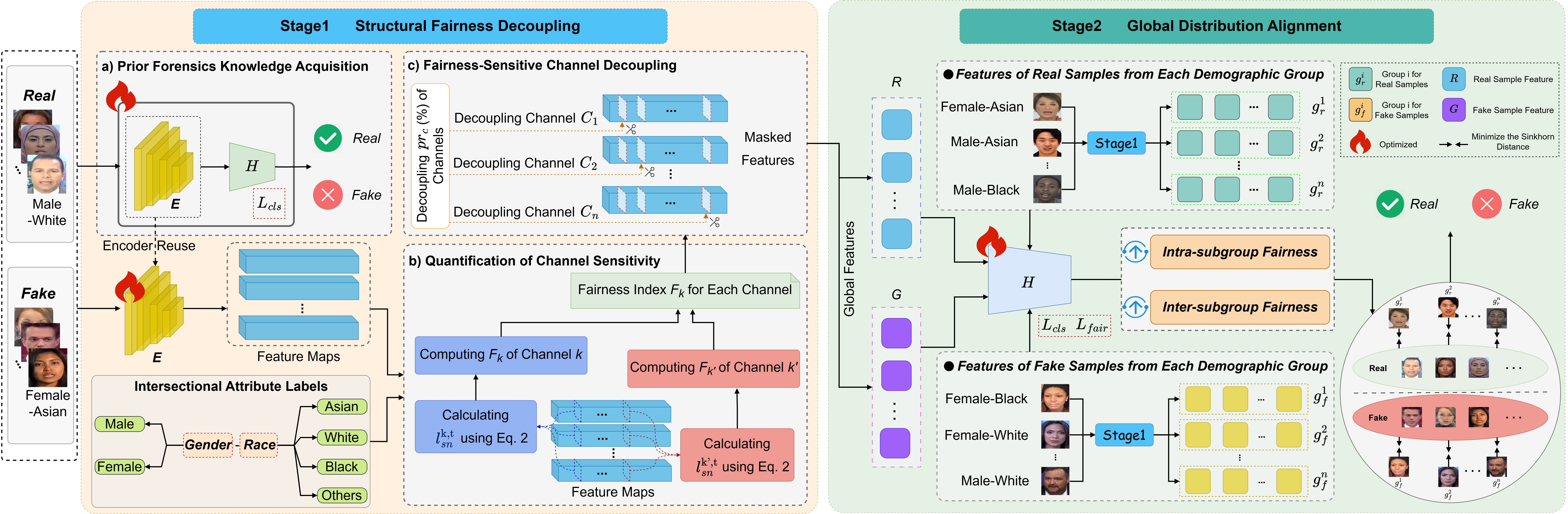}
  \vspace{-2mm}
  \caption{
    Overview of our proposed method: 
    (1) \textbf{Structural Fairness Decoupling}: dynamically identify and decouple channels that are highly correlated with sensitive attributes; 
    (2) \textbf{Global Distribution Alignment}: align prediction distributions of real and fake images within each sensitive subgroup with the global distributions. Specifically, this module adjusts the distributions of real images \( g_r^i \) and fake images \( g_f^i \) for different sensitive groups (\( i \in \mathcal{A} \)) to align with the global distributions of real images \( \mathcal{R} \) and fake images \( \mathcal{G} \).
  }
  \label{figure2}
  \vspace{-3mm}
\end{figure*}

\noindent
\textbf{Framework.} Fig.~\ref{figure2} presents an overview of the proposed framework, which comprises two stages: Structural Fairness Decoupling and Global Distribution Alignment. The purpose of the Structural Fairness Decoupling stage is to selectively decouple channels that are sensitive to attribute \(a_i\) from a specific layer of the model \(f_{\theta}\), thereby reducing the model's reliance on these sensitive attributes. In the Global Distribution Alignment stage, we align the distributions of real and fake images within each sensitive subgroup to the global distributions. This dual-path fairness optimization mechanism effectively mitigates the model’s prediction bias across different sensitive attribute groups. We will delve into each stage's specifics in the following sections.

\subsection{Structural Fairness Decoupling}
\vspace{-1mm}
To eliminate the implicit encoding of sensitive attributes in the feature representation layer of the model, we propose a fairness-sensitive channel decoupling strategy that reduces bias by optimizing the feature entanglement. 

\noindent
\textbf{Prior Forensics Knowledge Acquisition.} First, we adopt the Cross-Entropy Loss as the classification loss function to optimize the classification performance of the model. The classification loss is defined as:
\begin{equation} \label{eq:classification_loss}
L_{\text{cls}} = C(h(z_r^i), y_r^i) + C(h(z_f^i), y_f^i),
\end{equation}
where \(C(\cdot, \cdot)\) represents the Cross-Entropy Loss, \(h(\cdot)\) is the classification head,  and \(z_r^i\) and \(z_f^i\) are the feature representations of real and fake images, respectively. \(y_r^i\) and \(y_f^i\) are the true labels of the real and fake images, respectively. This loss function ensures that the model can accurately distinguish between real and fake images. Through this design, we provide support for subsequent fairness optimization.

\noindent
\textbf{Quantification of Channel Sensitivity.} The core idea of this module is that different convolutional channels in the deepfake detection model respond to sensitive attributes with significant heterogeneity. Some channels may introduce prediction bias by encoding local texture patterns strongly correlated with attributes such as skin color and gender (\textit{e.g.}, skin reflectance properties, facial contour geometrical features). To address this, we design a channel sensitivity evaluation criterion based on inter-class and intra-class feature similarity comparison, and dynamically quantify the fairness sensitivity of channels by regularizing the Soft Nearest Neighbor Loss (SNNL).

Specifically, for the \textit{t}-th bath, given the feature tensor \( M \in \mathbb{R}^{B \times C \times H \times W} \) (where \( B \) is the batch size, \( C \) is the number of channels, and \( H \times W \) represents the spatial dimensions) produced by the last convolution layer of the convolutional neural network, the feature map of each channel is denoted as \( M_k \in \mathbb{R}^{B \times H \times W} \). 
To evaluate the fairness sensitivity of channel \( k \), we define its sensitivity loss as:

{\footnotesize
\begin{equation} \label{eq:snnl_loss}
\small
\displaystyle l_{sn}^{\text{k,t}} = -\frac{1}{b} \sum\limits_{i=1}^{b} \log \left( \frac{\sum\limits_{x=1, x \neq i}^{b} \delta(a_i - a_x) \exp \left( \frac{-\| m_{k,i} - m_{k,x} \|^2}{T} \right)}{\sum\limits_{y=1, y \neq i}^{b} \exp \left( \frac{-\| m_{k,i} - m_{k,y} \|^2}{T} \right)} \right),
\end{equation}
}

where \( m_{k,i} \), \( m_{k,x} \) and \( m_{k,y} \) represent the feature representations of the \( i \)-th, \( x \)-th and \( y \)-th samples in channel \( k \), respectively. \( \delta(a_i - a_x) \) is Dirac’s Delta function, which equals 1 when \( a_i = a_x \) (\textit{i.e.}, the two samples belong to the same sensitive group) and 0 otherwise. \( T > 0 \) is the temperature parameter, used to control the sensitivity of feature similarity. \( b \) is the number of samples in the current \textit{t}-th batch.

To better observe the feature distribution ability of each channel across different sensitive groups, we compute the fairness index of channel \(k\) as 
\(F_k = \frac{1}{N_b}\sum_{t\in N_b}\bigl|l_{sn}^{k,t}\bigr|\), 
where \(N_b\) is the number of batches, and \( l_{sn}^{\text{k,t}} \) denotes the sensitivity loss of channel \( k \) in the \( t \)-th batch. A lower value of \( F_k \) indicates stronger discriminatory ability of channel \( k \) for sensitive attributes, thus potentially introducing more unfairness. By calculating the fairness index, we can optimize the fairness of the model by ensuring that channels provide clear and accurate information, leading to a more balanced model. 

\noindent
\textbf{Fairness-Sensitive Channel Decoupling.} Based on the computed fairness index \( F_k \) for each channel, we propose to decouple channels. Specifically, we sort all channels according to their fairness index \( F_k \) , and decouple the bottom \(pr_c\) percent, where \(pr_c\) is the channel decoupling ratio. In the ablation study, we will study its impact on  fairness. 

\subsection{Global Distribution Alignment}
\vspace{-1mm}
The improvement in fairness during the second stage is achieved by minimizing the distributional discrepancies across different groups.

Let \(D_{\{(x_I, a)\}|f}\) represent the distribution of feature given the model \(f\), and \( D_{\{(x_I, a)|a=\alpha\}|f} \) denote the distribution conditioned on a specific group \( \alpha \). The fairness objective is defined as:
\begin{equation}
\min_f \sum^A_\alpha d(D_{\{(x_I, a)\}|f} - D_{\{(x_I, a)|a=\alpha\}|f}),
\end{equation} 
where \( d \) is a distance function. Since the computation is intractable, we propose to align the empirical distributions \( D_{B|f} \) (estimated from batch \( B \)) and \( D_{B_a|f} \) (for group \( a \)), ensuring that the model’s predictions are invariant to sensitive attributes.
Specifically, for a sensitive attribute group \( a \in \mathcal{A} \), the predicted distributions of real and fake images are denoted as \( g_r^a \) and \( g_f^a \), respectively. The global distributions of real and fake images are denoted as \( \mathcal{R} \) and \( \mathcal{G} \). 
The proposed method aligns the predicted distributions of each sensitive group with the global distribution through optimal transport, while using mutual information to constrain the complexity of the transport plan. The transport cost function is formulated as:
\begin{equation} \label{eq:transport loss}
\small
\mathcal{L}_c^{\epsilon}(g^{a}_{r}, \mathcal{R}) = \min_{\substack{(X,Y) \\ X \sim g^{a}_{r} \\ Y \sim \mathcal{R}}} 
\left\{ \mathbb{E}_{(X,Y)} \left[ c(X,Y) \right] + \epsilon \cdot I(X;Y) \right\},
\end{equation}
where the mutual information term $I(X;Y)$ in the equation is defined as:
\begin{equation} \label{eq:mutual_info}
I(X;Y) = \text{KL}(\pi \parallel g_r^a \otimes \mathcal{R}).
\end{equation} 

In this formulation, $c(X, Y)$ represents the transport cost, which measures the difference between the sensitive attribute group and the global distribution. Equation \eqref{eq:mutual_info} shows that the KL divergence term quantifies the deviation between the joint distribution $\pi$ (describing the statistical dependence between $X$ and $Y$) and the product of marginal distributions $g_r^a \otimes \mathcal{R}$. The regularization coefficient $\epsilon > 0$ is a hyperparameter. The non-negativity of mutual information $I(X;Y)$ ensures it equals zero when $X$ and $Y$ are independent, thereby penalizing the model for violating the independence assumption between the sensitive attribute and the prediction.

Similarly, we define the transport cost for the predicted distribution of fake images as \( L_c^\epsilon(g_f^a, G) \), The total fairness loss is the average of the fairness losses across all sensitive attribute groups. Specifically, it is computed as:
\begin{equation} \label{eq:fairness_loss}
L_{\text{fair}} = \frac{1}{|A|} \sum_{a \in A} \left( L_c^\epsilon(g_r^a, \mathcal{R}) + L_c^\epsilon(g_f^a, \mathcal{G}) \right).
\end{equation}

During training, the model efficiently computes the above loss using the Sinkhorn-Knopp algorithm. For each input batch, the predicted probabilities for real and fake images from each sensitive group are first extracted. The empirical distributions are approximated using kernel density estimation, and a cost matrix is constructed based on the squared Euclidean distance between the predicted values of each group. The Gibbs kernel \( K = \exp(-C / \epsilon) \) is initialized. By alternately iterating between row normalization and column normalization vectors until convergence, the transport plan is obtained, and the transport cost is computed. The algorithm uses entropy regularization to reduce the complexity from the classical optimal transport problem's \( \mathcal{O}(n^3) \) to \( \mathcal{O}(n^2) \), significantly improving computational efficiency.

The final training objective for the second stage (global distribution alignment) combines the classification loss and fairness loss as follows:
\begin{equation} \label{eq:overall loss}
L_{\text{total}} = L_{\text{cls}} + \lambda L_{\text{fair}},
\end{equation}
where \( L_{\text{cls}} \) is the binary cross-entropy classification loss, and \( \lambda \) balances detection accuracy and fairness. This method reduces algorithmic bias by constraining the independence between sensitive groups and the global distribution, while maintaining high accuracy in deepfake detection.

\begin{table*}
\centering
\caption{Intra-domain evaluation on FF++ dataset. All methods are trained on FF++, tested on its test sub-datasets separated by sensitive attribute.}
\vspace{-1mm}
\label{table1}
\resizebox{0.8\linewidth}{!}{\scriptsize

\begin{tblr}{
  colspec = {c|c|c|c|c|c|c}, 
  rowsep = 0pt,
  cells  = {c, valign=m},
  cell{1}{1} = {r=2}{},  
  cell{1}{2} = {r=2}{},  
  cell{1}{3} = {r=2}{},  
  cell{1}{4} = {c=3}{halign=c, valign=m}, 
  cell{3}{1}  = {r=7}{},  
  cell{10}{1} = {r=7}{},  
  cell{17}{1} = {r=7}{},  
  row{9,16,23} = {bg=mygray},
  hline{1}  = {-}{},      
  hline{2}  = {-}{},      
  hline{3}  = {-}{},      
  hline{10} = {-}{},      
  hline{17} = {-}{},      
  hline{24} = {-}{},      
  vline{1} = {3-23}{wd=0pt},         
  vline{8} = {9,16,23}{wd=0pt},      
  vline{5} = {2-23}{wd=0pt}, 
  vline{6} = {2-23}{wd=0pt}, 
}

Attributes & Methods & Backbone & Fairness Metrics(\%) &  &  & Detection Metrics(\%) \\
          &         &          & $F_{FPR}$↓ & $F_{DP}$↓ & $es-AUC$↑ & $AUC$↑ \\
          
Gender       & Ori                            & Xception & 4.10  & 5.72  & 91.93 & 92.69 \\
             & DAG-FDD \textsubscript{\textcolor{blue}{WACV'24}}~\cite{ju2024improving}& Xception & 1.82  & \underline{4.65}  & 94.87 & 96.72 \\
             & DAW-FDD \textsubscript{\textcolor{blue}{WACV'24}}~\cite{ju2024improving}& Xception & 0.78  & 9.52  & 95.76 & 97.46 \\
             & PG-FDD \textsubscript{\textcolor{blue}{CVPR'24}}~\cite{lin2024preserving}& Xception & 0.62  & 4.74  & \underline{96.32} & \underline{97.66} \\
             & Fairadapter \textsubscript{\textcolor{blue}{ICASSP'25}}~\cite{ding2025fairadapter}& ViT-L/14 & 4.16 & 12.21 &67.85&71.50      \\
             & RSEF-FDD \textsubscript{\textcolor{blue}{arxiv'25}}~\cite{he2024redundant}& Xception &\underline{0.57}&8.55&94.91&97.09\\
             & Ours                            & Xception & \textbf{0.53}  & \textbf{3.61}  & \textbf{96.45} & \textbf{97.71} \\
             
Race         & Ori                            & Xception & 19.76 & \underline{4.74}  & 82.85 & 92.69 \\
             & DAG-FDD \textsubscript{\textcolor{blue}{WACV'24}}~\cite{ju2024improving}& Xception & \underline{5.48}  & 9.20  & 93.43 & 96.72 \\
             & DAW-FDD \textsubscript{\textcolor{blue}{WACV'24}}~\cite{ju2024improving}& Xception & \textbf{5.43}  & 14.69 & 94.15 & 97.46 \\
             & PG-FDD \textsubscript{\textcolor{blue}{CVPR'24}}~\cite{lin2024preserving}& Xception & 11.13 & 4.78  & \underline{94.52} & \underline{97.66} \\
             & Fairadapter \textsubscript{\textcolor{blue}{ICASSP'25}}\cite{ding2025fairadapter}& ViT-L/14 &43.22&20.39&56.03&71.50\\
             & RSEF-FDD \textsubscript{\textcolor{blue}{arxiv'25}}~\cite{he2024redundant}& Xception &8.39&5.28&93.60&97.09\\
             & Ours                            & Xception & 9.29  & \textbf{4.35}  & \textbf{94.86} & \textbf{97.71} \\
             
Intersection & Ori                            & Xception & 36.03 & 14.64 & 74.43 & 92.69 \\
             & DAG-FDD \textsubscript{\textcolor{blue}{WACV'24}}~\cite{ju2024improving}& Xception & 24.08 & 26.25 & 85.80 & 96.72 \\
             & DAW-FDD \textsubscript{\textcolor{blue}{WACV'24}}~\cite{ju2024improving}& Xception & \underline{14.36} & 26.13 & 86.74 & 97.46 \\
             & PG-FDD \textsubscript{\textcolor{blue}{CVPR'24}}~\cite{lin2024preserving}& Xception & \textbf{9.19}  & \underline{13.39} & \underline{86.83} & \underline{97.66} \\
             & Fairadapter \textsubscript{\textcolor{blue}{ICASSP'25}}~\cite{ding2025fairadapter}& ViT-L/14 &86.91&42.44&45.98&71.50\\
             & RSEF-FDD \textsubscript{\textcolor{blue}{arxiv'25}}~\cite{he2024redundant}& Xception &23.64&21.74&85.77&97.09\\
             & Ours                            & Xception & 20.18 & \textbf{9.47}  & \textbf{86.91} & \textbf{97.71} \\
\end{tblr}
}
\vspace{-1mm}
\end{table*}

\section{Experiment}
\vspace{-1mm}
\subsection{Experimental Settings}
\vspace{-1mm}
\textbf{Datasets.} To evaluate the fairness generalization capability of the proposed method, we trained the model on the widely used FaceForensics++ (FF++) dataset~\cite{rossler2019faceforensics++} and tested it on four benchmark datasets: FF++, DeepFake Detection Challenge (DFDC)~\cite{deepfake-detection-challenge}, DeepFakeDetection (DFD)~\cite{r12}, and Celeb-DF~\cite{li2020celeb}. To evaluate the fairness attribute, we divided the test dataset into intra-group evaluations focusing separately on gender and race, and inter-group evaluations combining both attributes (\textit{e.g.}, Male-White, Female-Asian). Details of each annotated dataset are in Appendix~\ref{sec:experimental_settings}.

\noindent
\textbf{Evaluation Metrics.} For detection comparison, the Area Under Curve (AUC) is used to benchmark our approach against previous works, which aligns with the detection evaluation approach adopted in precedent works~\cite{lin2024preserving, ju2024improving}. Regarding fairness, we use three distinct fairness metrics to evaluate the effectiveness of our proposed method. Specifically, we report the Equal False Positive Rate (\(F_{FPR}\)), Demographic Parity (\(F_{DP}\)) and \(es-AUC\).

{\footnotesize
\begin{equation}
\begin{aligned}\small
F_{FPR} := \sum_{J_j \in \mathcal{J}} 
  \Bigg| 
  &\frac{\sum_{j=1}^{n} \mathbb{I}[Y_j = 1, D_j = J_j, Y_j' = 0]}
         {\sum_{j=1}^{n} \mathbb{I}[D_j = J_j, Y_j' = 0]} 
   - {}\\
  &\frac{\sum_{j=1}^{n} \mathbb{I}[Y_j = 1, Y_j' = 0]}
         {\sum_{j=1}^{n} \mathbb{I}[Y_j' = 0]} 
  \Bigg|.
\end{aligned}
\end{equation}
\begin{equation}
\begin{aligned}\small
F_{DP} := \max_{k \in \{0,1\}} 
\Bigg\{ 
  &\max_{J_j \in \mathcal{J}} 
    \frac{\sum_{j=1}^{n} \mathbb{I}[\hat{Y}_j = k, D_j = J_j]}
         {\sum_{j=1}^{n} \mathbb{I}[D_j = J_j]}
  - {}\\
  &\min_{J_j' \in \mathcal{J}} 
    \frac{\sum_{j=1}^{n} \mathbb{I}[\hat{Y}_j = k, D_j = J_j']}
         {\sum_{j=1}^{n} \mathbb{I}[D_j = J_j']}
\Bigg\}.
\end{aligned}
\end{equation}
\begin{equation}\small
es-AUC := 
\frac{\text{Overall AUC}}
     {1 + \sum_{j=1}^{N} 
     \lvert \text{Overall AUC} - \text{AUC}_j \rvert}.
\end{equation}
}

Where \( D \) is the demographic variable, \( \mathcal{J} \) is the set of subgroups with each subgroup \( J_j \in \mathcal{J} \). \( F_{\text{FPR}} \) measures the disparity in False Positive Rate (FPR) across different groups compared to the overall population. \( F_{\text{DP}} \) measures the maximum difference in prediction rates across all demographic groups. \(es-AUC\) measures the overall fairness-consistent detection performance by penalizing disparities between the overall AUC and group-specific AUCs. The experimental results presented in all tables are used \(\uparrow\) to indicate that higher values are better, and \(\downarrow\) that lower values are better. By default, the best results are shown in \textbf{bold}, and the second-best results are indicated with \underline{underline}.

\noindent
\textbf{Baseline Methods.} We consider two widely used CNN architectures to validate the effectiveness of our proposed method (\textit{i.e.}, Xception~\cite{chollet2017xception}, ResNet-50~\cite{he2016deep}). We compare our method against the latest fairness method DAG-FDD~\cite{ju2024improving}, DAW-FDD~\cite{ju2024improving}, PG-FDD~\cite{lin2024preserving}, Fairadapter~\cite{ding2025fairadapter} and RSEF-FDD~\cite{he2024redundant} in deepfake detection. The compared method also includes `Ori' (a backbone with cross-entropy loss).

\noindent
\textbf{Implementation Details.} All experiments are based on PyTorch and trained with two NVIDIA RTX 4090. For training, we fix the batch size 64, epochs 50, use SGD optimizer with learning rate \(\beta\) = 1 \(\times\) \(10^{-3}\). The \(\epsilon\) in Eq.~\ref{eq:transport loss} is set to \(5 \times 10^{-4}\). For the overall loss, we set the \(\lambda\) in Eq.~\ref{eq:overall loss} as 0.005. 

\subsection{Experimental Results}

\begin{table*}[t]
\centering
\caption{Evaluation of different methods for improving fairness and detection generalization across cross-domain datasets (DFDC, Celeb-DF, and DFD).}
\label{table2}
\resizebox{\textwidth}{!}{
\begin{tblr}{
  colspec = {c|c|c|c c c|c c c|c c c|c},
  rowsep = 0pt,
  cells  = {c, valign=m},
  cell{1}{1} = {r=3}{},
  cell{1}{2} = {r=3}{},
  cell{1}{3} = {r=3}{},
  cell{1}{4}  = {c=9}{halign=c, valign=m},
  cell{2}{4}  = {c=3}{},
  cell{2}{7}  = {c=3}{},
  cell{2}{10} = {c=3}{},
  cell{4}{1}  = {r=7}{},
  cell{11}{1} = {r=7}{},
  cell{18}{1} = {r=7}{},
  row{10,17,24} = {bg=mygray},
  hline{1,25} = {-}{0.08em},
  hline{2,3}  = {1-13}{},
  hline{4,11,18} = {1-13}{},
}
Datasets & Methods & Backbone & Fairness Metrics(\%) &  &  &  &  &  &  &  &  & {\small Detection Metrics(\%)} \\
         &         &          & Gender &  &  & Race &  &  & Intersection &  &  & {\small Overall} \\
         &         &          & \(F_{FPR}\)↓ & \(F_{DP}\)↓ & \(es-AUC\)↑ & \(F_{FPR}\)↓ & \(F_{DP}\)↓ & \(es-AUC\)↑ & \(F_{FPR}\)↓ & \(F_{DP}\)↓ & \(es-AUC\)↑ & \(AUC\)↑ \\
DFDC     & Ori     & Xception & 8.67  & 6.70 & 54.56 & 19.41 & 7.99 & 47.02 & 53.42 & 17.34 & 36.96 & 56.13 \\
         & DAG-FDD \textsubscript{\textcolor{blue}{WACV'24}}~\cite{ju2024improving} & Xception & 5.29 & 6.68 & 57.41 & \textbf{9.26} & 12.51 & 45.91 & 45.30 & 12.92 & 36.42 & 59.13 \\
         & DAW-FDD \textsubscript{\textcolor{blue}{WACV'24}}~\cite{ju2024improving} & Xception & 3.60 & \underline{3.67} & 55.88 & 22.36 & 9.72 & 47.77 & 46.80 & 12.69 & 36.63 & 56.90 \\
         & PG-FDD \textsubscript{\textcolor{blue}{CVPR'24}}~\cite{lin2024preserving} & Xception & 2.35 & \textbf{2.57} & 57.71 & \underline{10.10} & 11.49 & 50.53 & \textbf{32.16} & 17.71 & 37.42 & \underline{59.64} \\
         & Fairadapter \textsubscript{\textcolor{blue}{ICASSP'25}}~\cite{ding2025fairadapter}& ViT-L/14 & 2.39 & 4.57 & 55.88 & 32.49 & 26.77 & \underline{52.27} & 74.10 & 31.53 & 31.39 & 59.25 \\
         & RSEF-FDD \textsubscript{\textcolor{blue}{arxiv'25}}~\cite{he2024redundant}& Xception & \underline{2.09} & 3.83 & \underline{57.90} & 12.24 & \underline{7.74} & 50.27 & 38.89 & \underline{12.18} & \underline{37.72} & 59.08 \\
         & Ours        & Xception & \textbf{1.76} & \underline{3.67} & \textbf{57.94} & 17.93 & \textbf{7.58} & \textbf{52.33} & \underline{37.37} & \textbf{11.92} & \textbf{39.03} & \textbf{61.86} \\
Celeb-DF & Ori         & Xception & 6.93 & 20.23 & 60.94 & 23.44 & 17.89 & 61.21 & 27.38 & 18.72 & 61.22 & 69.18 \\
         & DAG-FDD \textsubscript{\textcolor{blue}{WACV'24}}~\cite{ju2024improving} & Xception & 8.70 & 14.25 & 66.23 & 19.44 & 13.26 & 62.61 & \underline{23.60} & 13.80 & \underline{63.08} & 72.29 \\
         & DAW-FDD \textsubscript{\textcolor{blue}{WACV'24}}~\cite{ju2024improving} & Xception & 8.71 & \underline{12.73} & 68.31 & \textbf{14.76} & \underline{7.80} & 66.47 & 24.75 & 13.78 & 59.24 & 74.09 \\
         & PG-FDD \textsubscript{\textcolor{blue}{CVPR'24}}~\cite{lin2024preserving} & Xception & 7.95 & 16.29 & \underline{69.47} & 23.99 & 12.47 & 67.78 & 23.71 & \underline{13.30} & 62.63 & 74.24 \\
         & Fairadapter \textsubscript{\textcolor{blue}{ICASSP'25}}~\cite{ding2025fairadapter}& ViT-L/14 & 8.59 & 26.84 & 56.84 & 46.14 & 36.35 & 58.34 & 47.03 & 35.69 & 54.77 & 64.94 \\
         & RSEF-FDD \textsubscript{\textcolor{blue}{arxiv'25}}~\cite{he2024redundant}& Xception & \textbf{1.94} & 35.63 & 68.92 & 17.40 & 27.48 & \underline{67.80} & 24.17 & 27.73 & 62.56 & \underline{74.36} \\
         & Ours        & Xception & \underline{6.41} & \textbf{10.81} & \textbf{69.68} & \underline{16.60} & \textbf{7.29} & \textbf{69.96} & \textbf{23.32} & \textbf{12.01} & \textbf{63.27} & \textbf{76.75} \\
DFD      & Ori         & Xception & 9.45 & 6.63 & 72.68 & 7.75 & 22.31 & 69.07 & 35.06 & 25.53 & 59.20 & 74.32 \\
         & DAG-FDD \textsubscript{\textcolor{blue}{WACV'24}}~\cite{ju2024improving} & Xception & 5.41 & 5.60 & 75.87 & \underline{2.56} & 21.34 & 70.14 & 33.00 & 23.02 & 61.09 & 76.29 \\
         & DAW-FDD \textsubscript{\textcolor{blue}{WACV'24}}~\cite{ju2024improving} & Xception & 5.30 & 10.63 & 71.68 & 4.32 & 20.16 & 63.24 & 33.81 & 24.72 & 56.01 & 73.71 \\
         & PG-FDD \textsubscript{\textcolor{blue}{CVPR'24}}~\cite{lin2024preserving} & Xception & \underline{5.05} & \textbf{1.86} & \underline{76.37} & 5.79 & 19.31 & \underline{74.81} & \underline{28.00} & 23.93 & 66.32 & \underline{80.70} \\
         & Fairadapter \textsubscript{\textcolor{blue}{ICASSP'25}}~\cite{ding2025fairadapter}& ViT-L/14 & 6.32 & 7.55 & 63.28 & 11.29 & \textbf{12.28} & 56.28 & 29.56 & 33.11 & 48.63 & 68.12 \\
         &RSEF-FDD \textsubscript{\textcolor{blue}{arxiv'25}}~\cite{he2024redundant}& Xception & 18.50 & 6.11 & 75.97 & 19.45 & 19.40 & 72.65 & 35.34 & \textbf{10.97} & \underline{67.24} & 80.54 \\
         & Ours        & Xception & \textbf{4.72} & \underline{5.49} & \textbf{78.98} & \textbf{2.09} & \underline{18.66} & \textbf{77.44} & \textbf{27.32} & \underline{22.81} & \textbf{69.73} & \textbf{81.46} \\
\end{tblr}
}
\end{table*}

\begin{table*}[t]
\centering
\caption{Evaluation of fairness adaptation with different backbone across intra-domain (FF++) and cross-domain (DFDC, Celeb-DF, DFD) datasets.}
\label{table3}
\resizebox{\textwidth}{!}{
\begin{tblr}{
  colspec = {c|c|c|c c c|c c c|c c c|c},
  rowsep = 0pt,
  cells  = {c, valign=m},
  cell{1}{1} = {r=3}{},    
  cell{1}{2} = {r=3}{},    
  cell{1}{3} = {r=3}{},    
  cell{1}{4}  = {c=9}{halign=c, valign=m}, 
  cell{2}{4}  = {c=3}{},   
  cell{2}{7}  = {c=3}{},   
  cell{2}{10} = {c=3}{},   
  cell{4}{1}  = {r=5}{},   
  cell{9}{1}  = {r=5}{},   
  cell{14}{1} = {r=5}{},   
  cell{19}{1} = {r=5}{},   
  row{8,13,18,23} = {bg=mygray},
  hline{1,24} = {-}{0.08em},      
  hline{2,3}  = {1-13}{},         
  hline{4,9,14,19} = {1-13}{},    
}
Datasets & Methods & Backbone & Fairness Metrics(\%) &  &  &  &  &  &  &  &  & {\small Detection Metrics(\%)} \\
         &         &          & Gender &  &  & Race &  &  & Intersection &  &  & {\small Overall} \\
         &         &          & \(F_{FPR}\)↓ & \(F_{DP}\)↓ & \(es-AUC\)↑ & \(F_{FPR}\)↓ & \(F_{DP}\)↓ & \(es-AUC\)↑ & \(F_{FPR}\)↓ & \(F_{DP}\)↓ & \(es-AUC\)↑ & \(AUC\)↑ \\
FF++     & Ori     & ResNet-50 & 8.90 & 9.27 & 90.40 & 17.49 & 7.04 & 89.17 & 38.26 & 15.43 & 76.14 & 94.29 \\
         & DAG-FDD \textsubscript{\textcolor{blue}{WACV'24}}~\cite{ju2024improving} & ResNet-50 & \textbf{1.92} & 6.02 & 91.09 & \textbf{6.38} & 3.98 & 91.39 & 32.58 & 11.06 & 80.64 & \underline{94.74} \\
         & DAW-FDD \textsubscript{\textcolor{blue}{WACV'24}}~\cite{ju2024improving} & ResNet-50 & 3.25 & 5.29 & 90.13 & 12.46 & 3.29 & 90.92 & \underline{31.07} & 11.03 & 80.98 & 93.96 \\
         & PG-FDD \textsubscript{\textcolor{blue}{CVPR'24}}~\cite{lin2024preserving} & ResNet-50 & \underline{3.05} & \textbf{3.06} & \underline{93.57} & \underline{10.89} & \underline{2.83} & \underline{91.53} & \textbf{22.10} & \underline{9.69} & \underline{81.58} & 94.46 \\
         & Ours    & ResNet-50 & 3.94 & \underline{5.14} & \textbf{94.46} & 18.67 & \textbf{2.43} & \textbf{92.69} & 34.30 & \textbf{8.61} & \textbf{82.87} & \textbf{95.53} \\
DFDC     & Ori     & ResNet-50 & 12.55 & 11.06 & 54.96 & 17.34 & 8.38 & 48.30 & 48.10 & 16.30 & \underline{35.76} & 57.83 \\
         & DAG-FDD \textsubscript{\textcolor{blue}{WACV'24}}~\cite{ju2024improving} & ResNet-50 & 6.57 & 4.87 & \underline{57.63} & 15.96 & \underline{2.95} & \underline{50.31} & 44.10 & 11.25 & 33.41 & 60.29 \\
         & DAW-FDD \textsubscript{\textcolor{blue}{WACV'24}}~\cite{ju2024improving} & ResNet-50 & 9.12 & 7.14 & 57.22 & 14.60 & \textbf{2.81} & 48.94 & 43.31 & \underline{9.55} & 34.71 & 59.78 \\
         & PG-FDD \textsubscript{\textcolor{blue}{CVPR'24}}~\cite{lin2024preserving} & ResNet-50 & \underline{5.78} & \underline{4.65} & \textbf{58.34} & \underline{12.68} & 5.23 & 49.33 & \textbf{31.97} & \textbf{8.54} & 34.85 & \underline{61.83} \\
         & Ours    & ResNet-50 & \textbf{3.97} & \textbf{3.61} & 56.96 & \textbf{11.94} & 5.15 & \textbf{50.99 } & \underline{32.14} & 10.27 & \textbf{37.55 } & \textbf{62.54} \\
Celeb-DF & Ori     & ResNet-50 & 12.26 & 22.12 & 61.11 & 18.55 & 22.43 & 59.82 & 24.84 & 23.18 & 56.09 & 71.65 \\
         & DAG-FDD \textsubscript{\textcolor{blue}{WACV'24}}~\cite{ju2024improving} & ResNet-50 & 8.22 & 20.45 & 67.01 & 23.07 & 18.50 & 68.07 & \underline{20.29} & 19.33 & 66.76 & 75.77 \\
         & DAW-FDD \textsubscript{\textcolor{blue}{WACV'24}}~\cite{ju2024improving} & ResNet-50 & 9.03 & 20.85 & 68.33 & 18.6 & 17.65 & 69.14 & 23.84 & \underline{18.55} & \textbf{68.74} & \underline{76.51} \\
         & PG-FDD \textsubscript{\textcolor{blue}{CVPR'24}}~\cite{lin2024preserving} & ResNet-50 & \underline{6.98} & \underline{16.92} & \underline{68.49} & \underline{16.74} & \underline{16.50} & \underline{70.35} & 29.34 & 18.72 & 66.26 & 76.10 \\
         & Ours    & ResNet-50 & \textbf{4.53} & \textbf{12.00} & \textbf{68.98} & \textbf{6.76 } & \textbf{8.16 } & \textbf{71.03} & \textbf{5.56} & \textbf{9.11} & \underline{68.25} & \textbf{77.35} \\
DFD      & Ori     & ResNet-50 & 10.95 & 9.08 & 70.33 & 4.66 & 12.57 & 66.68 & 33.46 & 16.84 & 56.00 & 71.31 \\
         & DAG-FDD \textsubscript{\textcolor{blue}{WACV'24}}~\cite{ju2024improving} & ResNet-50 & 3.62 & 4.19 & 73.74 & \underline{3.86} & 13.54 & 70.33 & 25.22 & 14.84 & 57.19 & 73.97 \\
         & DAW-FDD \textsubscript{\textcolor{blue}{WACV'24}}~\cite{ju2024improving} & ResNet-50 & 4.81 & 3.90 & 72.12 & 7.36 & \underline{3.21} & 70.57 & 33.45 & 14.30 & 54.52 & 73.58 \\
         & PG-FDD \textsubscript{\textcolor{blue}{CVPR'24}}~\cite{lin2024preserving} & ResNet-50 & \underline{3.57} & \underline{1.82} & \underline{76.14} & \textbf{2.62} & 7.81 & \textbf{72.16} & \underline{23.15} & \underline{11.01} & \underline{62.65} & \underline{77.47} \\
         & Ours    & ResNet-50 & \textbf{3.12} & \textbf{1.76 } & \textbf{77.68} & 6.51 & \textbf{3.03 } & \underline{70.93} & \textbf{20.99} & \textbf{4.85} & \textbf{66.19} & \textbf{78.63  } \\
\end{tblr}
}
\end{table*}

\subsubsection{Performance of Intra-Domain}
As shown in Table~\ref{table1}, our method outperforms others in terms of fairness while achieving the best detection results. Whether in intra-group or inter-group evaluations, our method excels in most fairness metrics, demonstrating that the dual-mechanism collaborative optimization strategy not only mitigates bias in individual sensitive attributes but also effectively suppresses the complex bias accumulation caused by nonlinear interactions between attributes.

\subsubsection{Performance of Cross-Domain}
Table~\ref{table2} reports cross-domain fairness and detection results. Compared with other methods, our approach demonstrates superior fairness generalization while simultaneously achieving state-of-the-art detection performance. Notably, on the Celeb-DF dataset our method attains the best results on the intersection attribute and exhibits stronger generalization on individual attributes as well. By contrast, DAG-FDD and DAW-FDD show limited generalization and, in certain cross-domain settings, underperform the Ori baseline. Fairadapter, which is designed to improve fairness for AI-generated image detectors, performs poorly in the deepfake setting. PG-FDD and RSEF-FDD, though competitive and often stronger on fairness generalization, incur a drop in detection accuracy. Overall, our method outperforms all competitors on most fairness metrics and achieves the best results in both fairness generalization and AUC.

\subsubsection{Performance of Different Backbone}
To evaluate the fairness generalization capability of our proposed method with respect to backbone network selection, we replaced the Xception backbone with ResNet-50~\cite{he2016deep}. As shown in Table~\ref{table3}, our method achieves similarly superior results across different backbone. These findings demonstrate that the proposed approach is not constrained by the choice of backbone and remains effective and applicable under various backbone settings.

\begin{table*}
\centering
\caption{Ablation study results of the Xception backbone, comparing the original model (Ori), Global Distribution Alignment module (GDA), and GDA with Structural Fairness Decoupling (GDA + SFD).}
\label{table4}
\resizebox{\textwidth}{!}{
\begin{tabular}{@{}lccccccccccc@{}}
\toprule
\multicolumn{1}{c}{\multirow{3}{*}{Dataset}} &
\multicolumn{1}{c}{\multirow{3}{*}{Methods}} &
\multicolumn{9}{c}{Fairness Metrics(\%)} &
\multicolumn{1}{c}{\multirow{3}{*}{Detection Metrics(\%)}} \\
\cmidrule(lr){3-11}
 &  & \multicolumn{3}{c}{Gender} & \multicolumn{3}{c}{Race} & \multicolumn{3}{c}{Intersection} &  \\
\cmidrule(lr){3-5} \cmidrule(lr){6-8} \cmidrule(lr){9-11}

&  & $F_{FPR}$ \textdownarrow & $F_{DP}$ \textdownarrow & $es-AUC$ \textuparrow & 
$F_{FPR}$ \textdownarrow & $F_{DP}$ \textdownarrow & $es-AUC$ \textuparrow & 
$F_{FPR}$ \textdownarrow & $F_{DP}$ \textdownarrow & $es-AUC$ \textuparrow & $AUC$ \textuparrow \\

\midrule
\multirow{3}{*}{\parbox[c]{3cm}{\centering FF++}}
& Ori & 4.10 & 5.72 & 91.93 & 19.76 & 4.74 & 82.85 & 36.03 & 14.64 & 74.43 & 92.69 \\
& GDA & 3.91\textcolor{red}{(+0.19)} & 5.44\textcolor{red}{(+0.28)} & 96.11\textcolor{red}{(+4.18)} & 3.95\textcolor{red}{(+15.81)} & 3.23\textcolor{red}{(+1.51)} & 95.19\textcolor{red}{(+12.34)} & 26.85\textcolor{red}{(+9.18)} & 16.60\textcolor{blue}{(-1.96)} & 87.11\textcolor{red}{(+12.68)} & 97.22\textcolor{red}{(+4.53)} \\
& GDA+SFD & 0.53\textcolor{red}{(+3.38)} & 3.61\textcolor{red}{(+1.83)} & 96.45\textcolor{red}{(+0.34)} & 9.29\textcolor{blue}{(-5.34)} & 4.35\textcolor{blue}{(-1.12)} & 94.86\textcolor{blue}{(-0.33)} & 20.18\textcolor{red}{(+6.67)} & 9.47\textcolor{red}{(+7.13)} & 86.91\textcolor{blue}{(-0.2)} & 97.71\textcolor{red}{(+0.49)} \\
\midrule
\multirow{3}{*}{\parbox[c]{3cm}{\centering DFDC}}
& Ori & 8.67 & 6.70 & 54.56 & 19.41 & 7.99 & 47.02 & 53.42 & 17.34 & 36.96 & 56.13 \\
& GDA & 4.41\textcolor{red}{(+4.26)} & 4.36\textcolor{red}{(+2.34)} & 58.75\textcolor{red}{(+4.19)} & 20.77\textcolor{blue}{(-1.36)} & 11.6\textcolor{blue}{(-3.61)} & 43.69\textcolor{blue}{(-3.33)} & 46.28\textcolor{red}{(+7.14)} & 13.08\textcolor{red}{(+4.26)} & 32.45\textcolor{blue}{(-4.51)} & 60.91\textcolor{red}{(+4.78)} \\
& GDA+SFD & 1.76\textcolor{red}{(+2.65)} & 3.67\textcolor{red}{(+0.69)} & 57.94\textcolor{blue}{(-0.81)} & 17.93\textcolor{red}{(+2.84)} & 7.58\textcolor{red}{(+4.02)} & 52.33\textcolor{red}{(+8.64)} & 37.37\textcolor{red}{(+8.91)} & 11.92\textcolor{red}{(+1.16)} & 39.03\textcolor{red}{(+6.58)} & 61.86\textcolor{red}{(+0.95)} \\
\midrule
\multirow{3}{*}{\parbox[c]{3cm}{\centering Celeb-DF}}
& Ori & 6.93 & 20.23 & 60.94 & 23.44 & 17.89 & 61.21 & 27.38 & 18.72 & 61.22 & 69.18 \\
& GDA & 6.50\textcolor{red}{(+0.43)} & 18.18\textcolor{red}{(+2.05)} & 64.31\textcolor{red}{(+3.37)} & 22.96\textcolor{red}{(+0.48)} & 17.47\textcolor{red}{(+0.42)} & 64.92\textcolor{red}{(+3.71)} & 27.12\textcolor{red}{(+0.26)} & 18.12\textcolor{red}{(+0.6)} & 63.96\textcolor{red}{(+2.74)} & 72.89\textcolor{red}{(+3.71)} \\
& GDA+SFD & 6.41\textcolor{red}{(+0.09)} & 10.81\textcolor{red}{(+7.37)} & 69.68\textcolor{red}{(+5.37)} & 16.60\textcolor{blue}{(-6.36)} & 7.29\textcolor{red}{(+10.18)} & 69.96\textcolor{red}{(+5.04)} & 23.32\textcolor{red}{(+3.8)} & 12.01\textcolor{red}{(+6.11)} & 63.27\textcolor{blue}{(-0.69)} & 76.75\textcolor{red}{(+3.86)} \\
\midrule
\multirow{3}{*}{\parbox[c]{3cm}{\centering DFD}}
& Ori & 9.45 & 6.63 & 72.68 & 7.75 & 22.31 & 69.07 & 35.06 & 25.53 & 59.20 & 74.32 \\
& GDA & 8.89\textcolor{red}{(+0.56)} & 6.03\textcolor{red}{(+0.6)} & 74.47\textcolor{red}{(+1.79)} & 11.38\textcolor{blue}{(-3.63)} & 17.18\textcolor{red}{(+5.13)} & 75.24\textcolor{red}{(+6.17)} & 32.27\textcolor{red}{(+2.79)} & 20.46\textcolor{red}{(+5.07)} & 63.21\textcolor{red}{(+4.01)} & 77.78\textcolor{red}{(+3.46)} \\
& GDA+SFD & 4.72\textcolor{red}{(+4.17)} & 5.49\textcolor{red}{(+0.54)} & 78.98\textcolor{red}{(+4.51)} & 2.09\textcolor{red}{(+9.29)} & 18.66\textcolor{blue}{(-1.48)} & 77.44\textcolor{red}{(+2.2)} & 27.32\textcolor{red}{(+4.95)} & 22.81\textcolor{blue}{(-2.35)} & 69.73\textcolor{red}{(+6.52)} & 81.46\textcolor{red}{(+3.68)} \\
\bottomrule
\end{tabular}
}
\end{table*}

\begin{table*}
\centering
\caption{Ablation study results of the ResNet-50 backbone, comparing the original model (Ori), Global Distribution Alignment module (GDA), and GDA with Structural Fairness Decoupling (GDA $+$ SFD).}
\label{table5}
\resizebox{\textwidth}{!}{
\begin{tabular}{@{}lccccccccccc@{}}
\toprule
\multicolumn{1}{c}{\multirow{3}{*}{Dataset}} &
\multicolumn{1}{c}{\multirow{3}{*}{Methods}} &
\multicolumn{9}{c}{Fairness Metrics(\%)} &
\multicolumn{1}{c}{\multirow{3}{*}{Detection Metrics(\%)}} \\
\cmidrule(lr){3-11}
 &  & \multicolumn{3}{c}{Gender} & \multicolumn{3}{c}{Race} & \multicolumn{3}{c}{Intersection} &  \\
\cmidrule(lr){3-5} \cmidrule(lr){6-8} \cmidrule(lr){9-11}

&  & $F_{FPR}$ \textdownarrow & $F_{DP}$ \textdownarrow & $es-AUC$ \textuparrow & 
$F_{FPR}$ \textdownarrow & $F_{DP}$ \textdownarrow & $es-AUC$ \textuparrow & 
$F_{FPR}$ \textdownarrow & $F_{DP}$ \textdownarrow & $es-AUC$ \textuparrow & $AUC$ \textuparrow \\
\midrule

\multirow{3}{*}{\parbox[c]{3cm}{\centering FF++}} 
& Ori & 8.90 & 9.27 & 90.40 & 17.49 & 7.04 & 89.17 & 38.26 & 15.43 & 76.14 & 94.29 \\
& GDA & 3.86\textcolor{red}{(+5.04)} & 5.54\textcolor{red}{(+3.73)} & 91.70\textcolor{red}{(+1.3)} & 18.38\textcolor{blue}{(-0.89)} & 6.94\textcolor{red}{(+0.1)} & 89.72\textcolor{red}{(+0.55)} & 37.32\textcolor{red}{(+0.94)} & 11.10\textcolor{red}{(+4.33)} & 79.79\textcolor{red}{(+3.65)} & 94.75\textcolor{red}{(+0.46)} \\
& GDA+SFD & 3.94\textcolor{blue}{(-0.08)} & 5.14\textcolor{red}{(+0.4)} & 94.46\textcolor{red}{(+2.76)} & 18.67\textcolor{blue}{(-0.29)} & 2.43\textcolor{red}{(+4.51)} & 92.69\textcolor{red}{(+2.97)} & 34.30\textcolor{red}{(+3.02)} & 8.61\textcolor{red}{(+2.49)} & 82.87\textcolor{red}{(+3.08)} & 95.53\textcolor{red}{(+0.78)} \\
\midrule

\multirow{3}{*}{\parbox[c]{3cm}{\centering DFDC}}
& Ori & 12.55 & 11.06 & 54.96 & 17.34 & 8.38 & 48.30 & 48.10 & 16.30 & 35.76 & 57.83 \\
& GDA & 4.32\textcolor{red}{(+8.23)} & 2.01\textcolor{red}{(+9.05)} & 56.81\textcolor{red}{(+1.85)} & 17.25\textcolor{red}{(+0.09)} & 3.46\textcolor{red}{(+4.92)} & 46.16\textcolor{blue}{(-2.14)} & 45.50\textcolor{red}{(+2.60)} & 12.57\textcolor{red}{(+3.73)} & 34.12\textcolor{blue}{(-1.64)} & 60.58\textcolor{red}{(+2.75)} \\
& GDA+SFD & 3.97\textcolor{red}{(+0.35)} & 3.61\textcolor{blue}{(-1.60)} & 56.96\textcolor{red}{(0.15)} & 11.94\textcolor{red}{(+5.31)} & 5.15\textcolor{blue}{(-1.69)} & 50.99\textcolor{red}{(+4.83)} & 32.14\textcolor{red}{(+13.36)} & 10.27\textcolor{red}{(+2.30)} & 37.55\textcolor{red}{(+3.43)} & 62.54\textcolor{red}{(+1.96)} \\
\midrule

\multirow{3}{*}{\parbox[c]{3cm}{\centering Celeb-DF}}
& Ori & 12.26 & 22.12 & 61.11 & 18.55 & 22.43 & 59.82 & 24.84 & 23.18 & 56.09 & 71.65 \\
& GDA & 6.80\textcolor{red}{(+5.46)} & 14.07\textcolor{red}{(+8.05)} & 67.56\textcolor{red}{(+6.45)} & 21.51\textcolor{blue}{(-2.96)} & 10.73\textcolor{red}{(+11.70)} & 58.86\textcolor{blue}{(-0.96)} & 21.31\textcolor{red}{(+3.53)} & 11.18\textcolor{red}{(+12.00)} & 64.25\textcolor{red}{(+8.16)} & 72.40\textcolor{red}{(+0.75)} \\
& GDA+SFD & 4.53\textcolor{red}{(+2.27)} & 12.00\textcolor{red}{(+2.07)} & 68.98\textcolor{red}{(+1.42)} & 6.76\textcolor{red}{(+14.75)} & 8.16\textcolor{red}{(+2.57)} & 71.03\textcolor{red}{(+12.17)} & 5.56\textcolor{red}{(+15.75)} & 9.11\textcolor{red}{(+2.07)} & 68.25\textcolor{red}{(+4.00)} & 77.35\textcolor{red}{(+4.95)} \\
\midrule

\multirow{3}{*}{\parbox[c]{3cm}{\centering DFD}}
& Ori & 10.95 & 9.08 & 70.33 & 4.66 & 12.57 & 66.68 & 33.46 & 16.84 & 56.00 & 71.31 \\
& GDA & 7.46\textcolor{red}{(+3.49)} & 2.74\textcolor{red}{(+6.34)} & 70.77\textcolor{red}{(+0.44)} & 7.84\textcolor{blue}{(-3.18)} & 3.74\textcolor{red}{(+8.83)} & 67.84\textcolor{red}{(+1.16)} & 26.13\textcolor{red}{(+7.33)} & 9.20\textcolor{red}{(+7.64)} & 58.31\textcolor{red}{(+2.31)} & 70.99\textcolor{blue}{(-0.32)} \\
& GDA+SFD & 3.12\textcolor{red}{(+4.34)} & 1.76\textcolor{red}{(+0.98)} & 77.68\textcolor{red}{(+6.91)} & 6.51\textcolor{red}{(+1.33)} & 3.03\textcolor{red}{(+0.71)} & 70.93\textcolor{red}{(+3.09)} & 20.99\textcolor{red}{(+5.14)} & 4.85\textcolor{red}{(+4.35)} & 66.19\textcolor{red}{(+7.88)} & 78.63\textcolor{red}{(+7.64)} \\
\bottomrule
\end{tabular}
}

\end{table*}

\subsection{Robustness Study}
\begin{figure*}[!t]
\centering
\includegraphics[width=0.95\textwidth]{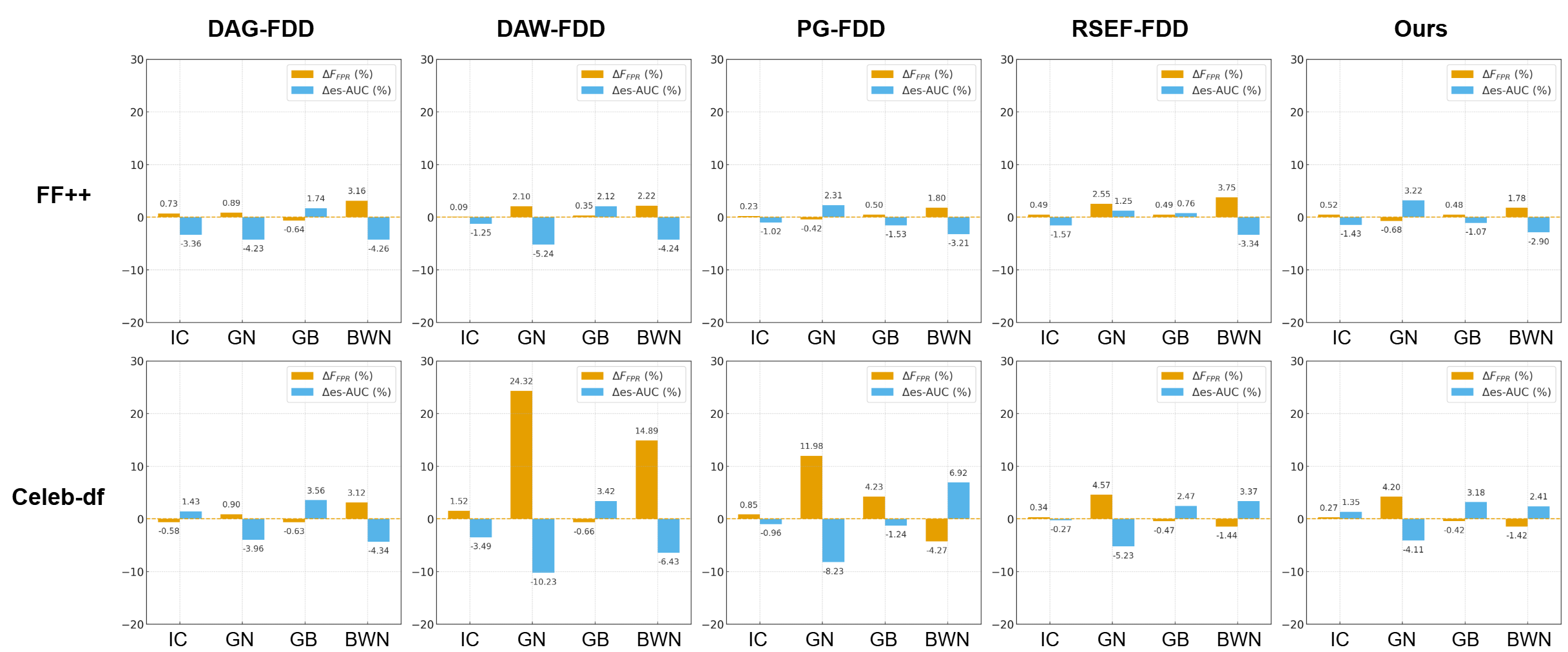}
\vspace{-1mm}
\caption{Robustness evaluation of proposed methods. We test the robustness of the proposed method against four disturbances of various intensities, and all models were trained on FF++.}
\label{fig:robustness}
\vspace{-1mm}
\end{figure*}

\vspace{-1mm}
Fig.~\ref{fig:robustness} presents the performance of all method under different degradation scenarios across FF++ and Celeb-df datasets, encompassing four common types of distortion: Image Compression (IC), Gaussian Noise (GN), Gaussian Blur (GB), and Block-Wise Noise (BWN). We introduce two additional metrics: \(\Delta F_{FPR}\) and \(\Delta {es-AUC}\). Experimental results show that our method exhibits comparable robustness to RSEF-FDD and outperforms other methods in terms of robustness. The robustness experiments conducted on the DFD and DFDC datasets are presented in the Appendix~\ref{sec:robustness_results}.

\subsection{Visualization Study}
\begin{figure*}[!t]
\centering
\includegraphics[width=0.95\textwidth]{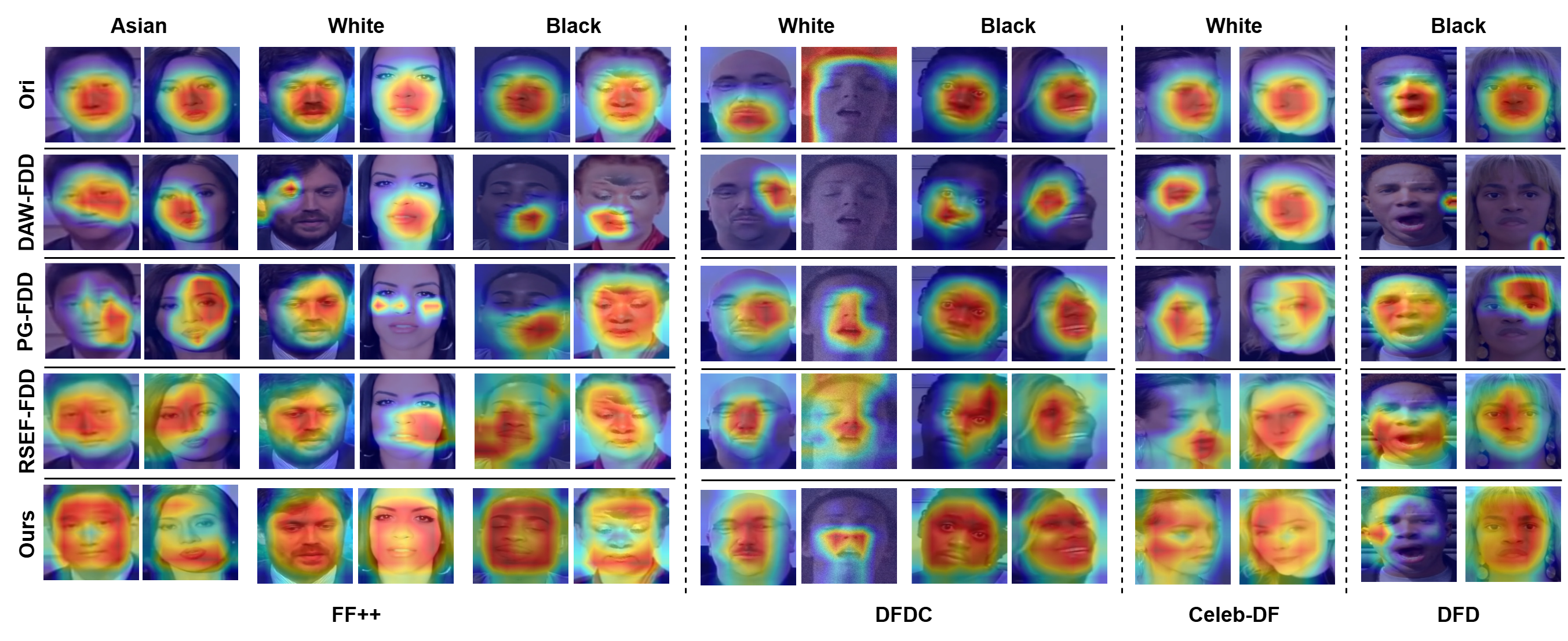}
\vspace{-2mm}
\caption{Grad-CAM visualization of ours and other methods on the intra-domain dataset (F++), and cross-domain datasets (DFDC, Celeb-DF, and DFD).}
\label{fig:visualization}
\vspace{-3mm}
\end{figure*}

\vspace{-1mm}
To better illustrate the effectiveness of our method, we visualize the Grad-CAM~\cite{selvaraju2017grad} results for Ori, DAW-FDD~\cite{ju2024improving}, PG-FDD~\cite{lin2024preserving}, RSEF-FDD~\cite{he2024redundant} and our method, as shown in Fig.~\ref{fig:visualization}. The Grad-CAM results reveal that Ori, in the absence of constraints, tends to overfit to small local regions or focus on background noise outside the facial area. Due to insufficient generalization, DAW-FDD performs similarly to Ori on cross-domain datasets. PG-FDD and RSEF-FDD boost fairness generalization and reduce local overfitting, but they are still vulnerable to non-facial background noise. In contrast, the activation maps of our method demonstrate that the model consistently attends to salient facial features across different datasets.

\subsection{Ablation Study}
\begin{figure}[htbp]
\centering
\includegraphics[width=0.98\linewidth]{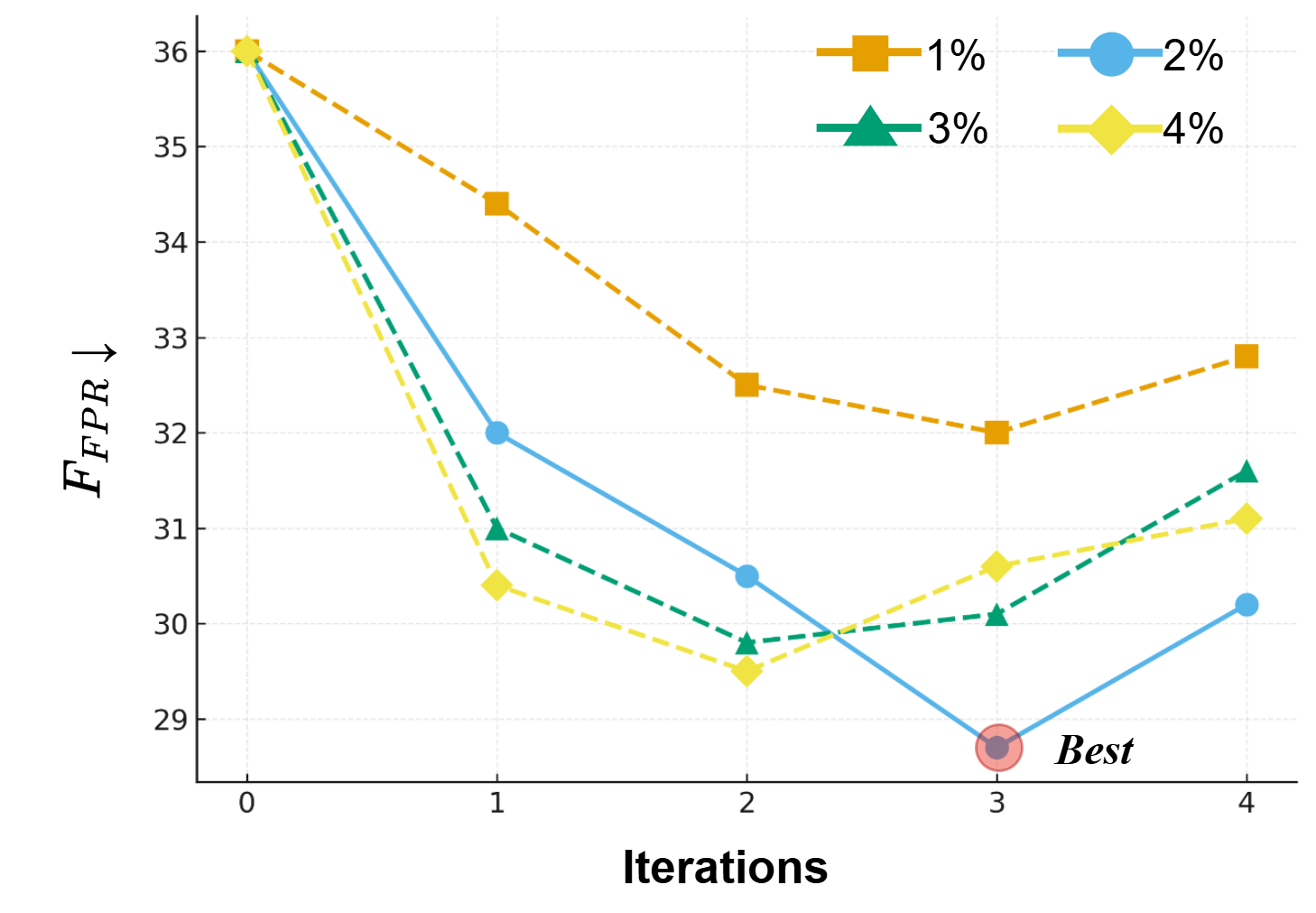}
\vspace{-1mm}
\caption{Analysis on the FF++ test set illustrating the impact of different decoupling iterations and decoupling ratios on the fairness performance (\(F_{FPR}\)) for Xception backnone.}
\label{figure4}
\vspace{-2mm}
\end{figure}

\vspace{-1mm}
To thoroughly analyze the effectiveness of our proposed modules, we conduct systematic ablation studies from both structural and parametric perspectives. As illustrated in Fig.~\ref{figure4}, we first investigate the influence of different decoupling ratios and iteration numbers within the structural fairness decoupling module on fairness performance. The results show that increasing the number of decoupling iterations initially improves the fairness performance (measured by \(F_{FPR}\)), but excessive decoupling leads to degradation. Empirically, the optimal trade-off is achieved when decoupling 2\% of channels during the third iteration, which results in the best balance between fairness and robustness. The fairness performance of the ResNet-50 backbone is analyzed in the Appendix~\ref{sec:ablation_results}.


Building upon this, Table~\ref{table4} and Table~\ref{table5} further reveal the collaborative optimization mechanism between the two key modules: Global Distribution Alignment (GDA) and Structural Fairness Decoupling (SFD). For instance, on the FF++ dataset, the gender prediction disparity (\(F_{\text{DP}}\)) of ResNet-50 improved from 9.27\% (Ori) to 5.54\% (GDA). With the inclusion of the SFD module, the gender \(F_{\text{FPR}}\) of Xception was further reduced to 0.53\% (↓87.1\%). In terms of detection accuracy, both models achieved significant AUC improvements (Xception: +5.02\%; ResNet-50: +1.24\%), indicating that the global optimization by GDA and the local fairness enhancement by SFD jointly suppress bias while preserving crucial forgery features.

\vspace{-1mm}
\section{Conclusion}
\vspace{-1mm}
This paper proposes a dual-mechanism collaborative optimization framework that combines structural fairness decoupling and global distribution alignment, effectively addressing the issue of fairness in deepfake detection. Experimental results demonstrate that the proposed method successfully reduces misclassification differences for sensitive attributes across multiple datasets, while maintaining excellent detection performance.


\section*{Acknowledgments}
This work was supported in part by the National Natural Science Foundation of China under Grant 62262041, and in part by the Jiangxi Provincial Natural Science Foundation under Grant 20232BAB202011.

{
    \small
    \bibliographystyle{ieeenat_fullname}
    \bibliography{main}

@String(ICASSP=	{ICASSP})

@String(AAAI = {AAAI})

@inproceedings{lin2024preserving,
  title={Preserving fairness generalization in deepfake detection},
  author={Lin, Li and He, Xinan and Ju, Yan and Wang, Xin and Ding, Feng and Hu, Shu},
  booktitle={Proceedings of the IEEE/CVF Conference on Computer Vision and Pattern Recognition},
  pages={16815--16825},
  year={2024}
}

@article{tolosana2020deepfakes,
  title={Deepfakes and beyond: A survey of face manipulation and fake detection},
  author={Tolosana, Ruben and Vera-Rodriguez, Ruben and Fierrez, Julian and Morales, Aythami and Ortega-Garcia, Javier},
  journal={Information Fusion},
  volume={64},
  pages={131--148},
  year={2020},
  publisher={Elsevier}
}

@book{citron2019deepfakes,
  title={How deepfakes undermine truth and threaten democracy},
  author={Citron, Danielle Keats},
  year={2019},
  publisher={TED}
}

@inproceedings{ju2024improving,
  title={Improving fairness in deepfake detection},
  author={Ju, Yan and Hu, Shu and Jia, Shan and Chen, George H and Lyu, Siwei},
  booktitle={Proceedings of the IEEE/CVF Winter Conference on Applications of Computer Vision},
  pages={4655--4665},
  year={2024}
}

@article{masood2023deepfakes,
  title={Deepfakes generation and detection: State-of-the-art, open challenges, countermeasures, and way forward},
  author={Masood, Momina and Nawaz, Mariam and Malik, Khalid Mahmood and Javed, Ali and Irtaza, Aun and Malik, Hafiz},
  journal={Applied intelligence},
  volume={53},
  number={4},
  pages={3974--4026},
  year={2023},
  publisher={Springer}
}

@misc{deepfake-detection-challenge,
    author = {benpflaum and Brian G and djdj and Irina Kofman and JE Tester and JLElliott and Joshua Metherd and Julia Elliott and Mozaic and Phil Culliton and Sohier Dane and Woo Kim},
    title = {Deepfake Detection Challenge},
    year = {2019},
    howpublished = {\url{https://kaggle.com/competitions/deepfake-detection-challenge}},
    note = {Kaggle}
}

@inproceedings{leibowicz2021deepfake,
  title={The deepfake detection dilemma: A multistakeholder exploration of adversarial dynamics in synthetic media},
  author={Leibowicz, Claire R and McGregor, Sean and Ovadya, Aviv},
  booktitle={Proceedings of the 2021 AAAI/ACM Conference on AI, Ethics, and Society},
  pages={736--744},
  year={2021}
}

@misc{r12,
  author       = {{Google} and {Jigsaw}},
  title        = {Deepfakes Dataset by Google \& Jigsaw},
  year         = {2019},
  howpublished = {\url{https://ai.googleblog.com/2019/09/contributing-data-to-deepfakedetection.html}},
}

@article{passos2024review,
  title={A review of deep learning-based approaches for deepfake content detection},
  author={Passos, Leandro A and Jodas, Danilo and Costa, Kelton AP and Souza J{\'u}nior, Luis A and Rodrigues, Douglas and Del Ser, Javier and Camacho, David and Papa, Jo{\~a}o Paulo},
  journal={Expert Systems},
  volume={41},
  number={8},
  pages={e13570},
  year={2024},
  publisher={Wiley Online Library}
}

@inproceedings{li2020celeb,
  title={Celeb-df: A large-scale challenging dataset for deepfake forensics},
  author={Li, Yuezun and Yang, Xin and Sun, Pu and Qi, Honggang and Lyu, Siwei},
  booktitle={Proceedings of the IEEE/CVF conference on computer vision and pattern recognition},
  pages={3207--3216},
  year={2020}
}

@article{albiero2021gendered,
  title={Gendered differences in face recognition accuracy explained by hairstyles, makeup, and facial morphology},
  author={Albiero, V{\'\i}tor and Zhang, Kai and King, Michael C and Bowyer, Kevin W},
  journal={IEEE Transactions on Information Forensics and Security},
  volume={17},
  pages={127--137},
  year={2021},
  publisher={IEEE}
}

@inproceedings{krishnan2020understanding,
  title={Understanding fairness of gender classification algorithms across gender-race groups},
  author={Krishnan, Anoop and Almadan, Ali and Rattani, Ajita},
  booktitle={2020 19th IEEE international conference on machine learning and applications (ICMLA)},
  pages={1028--1035},
  year={2020},
  organization={IEEE}
}

@inproceedings{ramachandran2021experimental,
  title={An experimental evaluation on deepfake detection using deep face recognition},
  author={Ramachandran, Sreeraj and Nadimpalli, Aakash Varma and Rattani, Ajita},
  booktitle={2021 International Carnahan Conference on Security Technology (ICCST)},
  pages={1--6},
  year={2021},
  organization={IEEE}
}

@article{guo2022robust,
  title={Robust attentive deep neural network for detecting gan-generated faces},
  author={Guo, Hui and Hu, Shu and Wang, Xin and Chang, Ming-Ching and Lyu, Siwei},
  journal={IEEE Access},
  volume={10},
  pages={32574--32583},
  year={2022},
  publisher={IEEE}
}

@article{pu2022learning,
  title={Learning a deep dual-level network for robust DeepFake detection},
  author={Pu, Wenbo and Hu, Jing and Wang, Xin and Li, Yuezun and Hu, Shu and Zhu, Bin and Song, Rui and Song, Qi and Wu, Xi and Lyu, Siwei},
  journal={Pattern Recognition},
  volume={130},
  pages={108832},
  year={2022},
  publisher={Elsevier}
}

@inproceedings{rossler2019faceforensics++,
  title={Faceforensics++: Learning to detect manipulated facial images},
  author={Rossler, Andreas and Cozzolino, Davide and Verdoliva, Luisa and Riess, Christian and Thies, Justus and Nie{\ss}ner, Matthias},
  booktitle={Proceedings of the IEEE/CVF international conference on computer vision},
  pages={1--11},
  year={2019}
}

@inproceedings{pu2022fairness,
  title={Fairness evaluation in deepfake detection models using metamorphic testing},
  author={Pu, Muxin and Kuan, Meng Yi and Lim, Nyee Thoang and Chong, Chun Yong and Lim, Mei Kuan},
  booktitle={Proceedings of the 7th international workshop on metamorphic testing},
  pages={7--14},
  year={2022}
}

@article{xu2024analyzing,
  title={Analyzing fairness in deepfake detection with massively annotated databases},
  author={Xu, Ying and Terh{\"o}rst, Philipp and Pedersen, Marius and Raja, Kiran},
  journal={IEEE Transactions on Technology and Society},
  volume={5},
  number={1},
  pages={93--106},
  year={2024},
  publisher={IEEE}
}

@article{trinh2021examination,
  title={An examination of fairness of ai models for deepfake detection},
  author={Trinh, Loc and Liu, Yan},
  journal={arXiv preprint arXiv:2105.00558},
  year={2021}
}

@article{hazirbas2021towards,
  title={Towards measuring fairness in ai: the casual conversations dataset},
  author={Hazirbas, Caner and Bitton, Joanna and Dolhansky, Brian and Pan, Jacqueline and Gordo, Albert and Ferrer, Cristian Canton},
  journal={IEEE Transactions on Biometrics, Behavior, and Identity Science},
  volume={4},
  number={3},
  pages={324--332},
  year={2021},
  publisher={IEEE}
}

@inproceedings{nadimpalli2022gbdf,
  title={GBDF: Gender balanced deepfake dataset towards fair deepfake detection},
  author={Nadimpalli, Aakash Varma and Rattani, Ajita},
  booktitle={International Conference on Pattern Recognition},
  pages={320--337},
  year={2022},
  organization={Springer}
}

@inproceedings{wang2022fairness,
  title={Fairness-aware adversarial perturbation towards bias mitigation for deployed deep models},
  author={Wang, Zhibo and Dong, Xiaowei and Xue, Henry and Zhang, Zhifei and Chiu, Weifeng and Wei, Tao and Ren, Kui},
  booktitle={Proceedings of the IEEE/CVF conference on computer vision and pattern recognition},
  pages={10379--10388},
  year={2022}
}

@article{little2022fairness,
  title={To the fairness frontier and beyond: Identifying, quantifying, and optimizing the fairness-accuracy pareto frontier},
  author={Little, Camille Olivia and Weylandt, Michael and Allen, Genevera I},
  journal={arXiv preprint arXiv:2206.00074},
  year={2022}
}

@inproceedings{roh2020fr,
  title={Fr-train: A mutual information-based approach to fair and robust training},
  author={Roh, Yuji and Lee, Kangwook and Whang, Steven and Suh, Changho},
  booktitle={International Conference on Machine Learning},
  pages={8147--8157},
  year={2020},
  organization={PMLR}
}

@inproceedings{sarhan2020fairness,
  title={Fairness by learning orthogonal disentangled representations},
  author={Sarhan, Mhd Hasan and Navab, Nassir and Eslami, Abouzar and Albarqouni, Shadi},
  booktitle={Computer Vision--ECCV 2020: 16th European Conference, Glasgow, UK, August 23--28, 2020, Proceedings, Part XXIX 16},
  pages={746--761},
  year={2020},
  organization={Springer}
}

@article{tifrea2023frappe,
  title={Frapp{\'e}: A group fairness framework for post-processing everything},
  author={Tifrea, Alexandru and Lahoti, Preethi and Packer, Ben and Halpern, Yoni and Beirami, Ahmad and Prost, Flavien},
  journal={arXiv preprint arXiv:2312.02592},
  year={2023}
}

@article{hardt2016equality,
  title={Equality of opportunity in supervised learning},
  author={Hardt, Moritz and Price, Eric and Srebro, Nati},
  journal={Advances in neural information processing systems},
  volume={29},
  year={2016}
}

@article{pleiss2017fairness,
  title={On fairness and calibration},
  author={Pleiss, Geoff and Raghavan, Manish and Wu, Felix and Kleinberg, Jon and Weinberger, Kilian Q},
  journal={Advances in neural information processing systems},
  volume={30},
  year={2017}
}

@article{ding2024disrupting,
  title={Disrupting Anti-Spoofing Systems by Images of Consistent Identity},
  author={Ding, Feng and Jiang, Zihan and Zhou, Yue and Xu, Jianfeng and Zhu, Guopu},
  journal={IEEE Signal Processing Letters},
  year={2024},
  publisher={IEEE}
}

@article{ye2024decoupling,
  title={Decoupling forgery semantics for generalizable deepfake detection},
  author={Ye, Wei and He, Xinan and Ding, Feng},
  journal={arXiv preprint arXiv:2406.09739},
  year={2024}
}

@article{he2024redundant,
  title={Redundant Semantic Environment Filling via Misleading-Learning for Fair Deepfake Detection},
  author={He, Xinan and Zhou, Yue and Hu, Shu and Li, Bin and Huang, Jiwu and Ding, Feng},
  journal={arXiv preprint arXiv:2405.15173},
  year={2024}
}

@article{fan2025generating,
  title={Generating Higher-Quality Anti-Forensics DeepFakes with Adversarial Sharpening Mask},
  author={Fan, Bing and Ding, Feng and Zhu, Guopu and Huang, Jiwu and Kwong, Sam and Atrey, Pradeep K and Lyu, Siwei},
  journal={ACM Transactions on Multimedia Computing, Communications and Applications},
  publisher={ACM New York, NY},
  year=[2025]
}

@article{ding2022securing,
  title={Securing facial bioinformation by eliminating adversarial perturbations},
  author={Ding, Feng and Fan, Bing and Shen, Zhangyi and Yu, Keping and Srivastava, Gautam and Dev, Kapal and Wan, Shaohua},
  journal={IEEE Transactions on Industrial Informatics},
  volume={19},
  number={5},
  pages={6682--6691},
  year={2022},
  publisher={IEEE}
}

@article{ding2021anti,
  title={Anti-forensics for face swapping videos via adversarial training},
  author={Ding, Feng and Zhu, Guopu and Li, Yingcan and Zhang, Xinpeng and Atrey, Pradeep K and Lyu, Siwei},
  journal={IEEE Transactions on Multimedia},
  volume={24},
  pages={3429--3441},
  year={2021},
  publisher={IEEE}
}

@inproceedings{selvaraju2017grad,
  title={Grad-cam: Visual explanations from deep networks via gradient-based localization},
  author={Selvaraju, Ramprasaath R and Cogswell, Michael and Das, Abhishek and Vedantam, Ramakrishna and Parikh, Devi and Batra, Dhruv},
  booktitle={Proceedings of the IEEE international conference on computer vision},
  pages={618--626},
  year={2017}
}

@inproceedings{chollet2017xception,
  title={Xception: Deep learning with depthwise separable convolutions},
  author={Chollet, Fran{\c{c}}ois},
  booktitle={Proceedings of the IEEE conference on computer vision and pattern recognition},
  pages={1251--1258},
  year={2017}
}

@inproceedings{he2016deep,
  title={Deep residual learning for image recognition},
  author={He, Kaiming and Zhang, Xiangyu and Ren, Shaoqing and Sun, Jian},
  booktitle={Proceedings of the IEEE conference on computer vision and pattern recognition},
  pages={770--778},
  year={2016}
}

@inproceedings{cheng2025fair,
  title={Fair Deepfake Detectors Can Generalize},
  author={Cheng, Harry and Liu, Ming-Hui and Guo, Yangyang and Wang, Tianyi and Nie, Liqiang and Kankanhalli, Mohan},
  booktitle={NeurIPS},
  year={2025}
}

@inproceedings{ding2025fairadapter,
  title={Fairadapter: Detecting ai-generated images with improved fairness},
  author={Ding, Feng and Zhang, Jun and He, Xinan and Xu, Jianfeng},
  booktitle={ICASSP 2025-2025 IEEE International Conference on Acoustics, Speech and Signal Processing (ICASSP)},
  pages={1--5},
  year={2025},
  organization={IEEE}
}

@incollection{lyu2022deepfake,
  title={Deepfake detection},
  author={Lyu, Siwei},
  booktitle={Multimedia forensics},
  pages={313--331},
  year={2022},
  publisher={Springer Singapore Singapore}
}

@article{wang2022deepfake,
  title={Deepfake noise investigation and detection},
  author={Wang, Tianyi and Liu, Ming and Cao, Wei and Chow, Kam Pui},
  journal={Forensic Science International: Digital Investigation},
  volume={42},
  pages={301395},
  year={2022},
  publisher={Elsevier}
}

@inproceedings{cao2022end,
  title={End-to-end reconstruction-classification learning for face forgery detection},
  author={Cao, Junyi and Ma, Chao and Yao, Taiping and Chen, Shen and Ding, Shouhong and Yang, Xiaokang},
  booktitle={Proceedings of the IEEE/CVF conference on computer vision and pattern recognition},
  pages={4113--4122},
  year={2022}
}

@article{he2025vlforgery,
  title={VLForgery Face Triad: Detection, Localization and Attribution via Multimodal Large Language Models},
  author={He, Xinan and Zhou, Yue and Fan, Bing and Li, Bin and Zhu, Guopu and Ding, Feng},
  journal={arXiv preprint arXiv:2503.06142},
  year={2025}
}

@article{zhou2025breaking,
  title={Breaking Latent Prior Bias in Detectors for Generalizable AIGC Image Detection},
  author={Zhou, Yue and He, Xinan and Lin, KaiQing and Fan, Bin and Ding, Feng and Li, Bin},
  journal={arXiv preprint arXiv:2506.00874},
  year={2025}
}

@inproceedings{liu2025thinking,
  title={Thinking racial bias in fair forgery detection: Models, datasets and evaluations},
  author={Liu, Decheng and Wang, Zongqi and Peng, Chunlei and Wang, Nannan and Hu, Ruimin and Gao, Xinbo},
  booktitle={Proceedings of the AAAI Conference on Artificial Intelligence},
  volume={39},
  number={5},
  pages={5379--5387},
  year={2025}
}

@article{zhou2026simplicity,
  title={Simplicity Prevails: The Emergence of Generalizable AIGI Detection in Visual Foundation Models},
  author={Zhou, Yue and He, Xinan and Lin, Kaiqing and Fan, Bing and Ding, Feng and Li, Bin},
  journal={arXiv preprint arXiv:2602.01738},
  year={2026}
}

@article{ding2026diffface,
  title={DiffFace-Edit: A Diffusion-Based Facial Dataset for Forgery-Semantic Driven Deepfake Detection Analysis},
  author={Ding, Feng and Yi, Wenhui and He, Xinan and Xiao, Mengyao and Xu, Jianfeng and Du, Jianqiang},
  journal={arXiv preprint arXiv:2601.13551},
  year={2026}
}
}

\clearpage
\appendix
\onecolumn    

\section*{\centering \Large Appendix for ``Decoupling Bias, Aligning Distributions: Synergistic Fairness Optimization for Deepfake Detection''}

\numberwithin{equation}{section}
\numberwithin{figure}{section}
\numberwithin{table}{section}

\renewcommand{\thesection}{{\Alph{section}}}                 
\renewcommand{\thesubsection}{\Alph{section}.\arabic{subsection}}  
\renewcommand{\thesubsubsection}{\Roman{section}.\arabic{subsection}.\arabic{subsubsection}}

\section{Additional Experimental Settings}
\label{sec:experimental_settings}
Table~\ref{tab:dataset_detail} reports the total numbers of training, validation, and test samples for each dataset, together with the sensitive attributes considered in our experiments. Training and validation are conducted exclusively on FF++ dataset.

\begin{table}[!htbp]
\caption{Sample usage for training and testing on the FF++, DFD, DFDC, and Celeb-DF datasets. `-' means not used.}
\label{tab:dataset_detail}
\centering
\resizebox{\linewidth}{!}{
\begin{tabular}{c|ccc|ccc}
\hline
\multirow{2}{*}{Dataset} & \multicolumn{3}{c|}{Samples} & \multicolumn{3}{c}{Sensitive Attributes} \\ \cline{2-7}
& Train & Validation & Test & Gender & Race & Intersection \\ \hline
FF++     & 76139 & 25386 & 25401 & Male, Female & Asian, Black, White, Others &
\begin{tabular}[c]{@{}c@{}}Male-Asian, Male-Black, Male-White, Male-Others\\ Female-Asian, Female-Black, Female-White, Female-Others\end{tabular} \\ \hline
DFD      & - & - & 9385  & Male, Female & Black, White, Others &
\begin{tabular}[c]{@{}c@{}}Male-Black, Male-White, Male-Others\\ Female-Black, Female-White, Female-Others\end{tabular} \\ \hline
DFDC     & - & - & 22857 & Male, Female & Asian, Black, White, Others &
\begin{tabular}[c]{@{}c@{}}Male-Asian, Male-Black, Male-White, Male-Others\\ Female-Asian, Female-Black, Female-White, Female-Others\end{tabular} \\ \hline
Celeb-DF & - & - & 28458 & Male, Female & Black, White, Others &
\begin{tabular}[c]{@{}c@{}}Male-Black, Male-White, Male-Others\\ Female-Black, Female-White, Female-Others\end{tabular} \\ \hline
\end{tabular}
}
\end{table}

\section{Additional Robustness Results}
\label{sec:robustness_results}
\begin{figure}[!htbp]
\centering
\includegraphics[width=\linewidth]{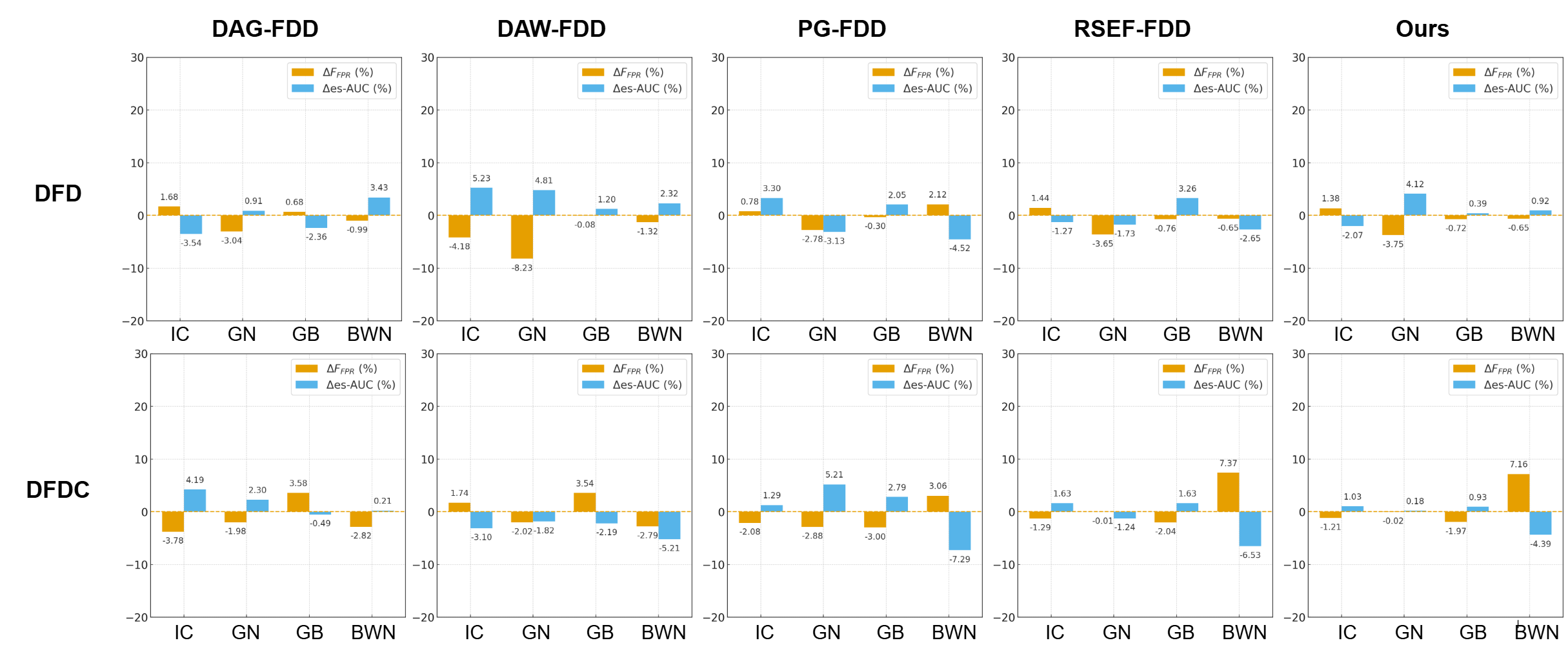}
\vspace{-1mm}
\caption{Additional robustness evaluation on the DFD and DFDC datasets. All methods are tested under four different types of perturbations.}
\label{fig:robustness2}
\vspace{-1mm}
\end{figure}

As shown in Fig.~\ref{fig:robustness2}, all methods are evaluated for robustness on the DFD and DFDC datasets under four types of perturbations: IC (Image Compression), GN (Gaussian Noise), GB (Gaussian Blur), and BWN (Block-wise Noise). The results show that our proposed method is more robust than the other baselines, with its performance remaining almost unchanged under most perturbations and even improving under GN perturbation on the DFD dataset.

\section{Additional Ablation Results}
\label{sec:ablation_results}
\begin{figure}[htbp]
\centering
\includegraphics[width=0.5\linewidth]{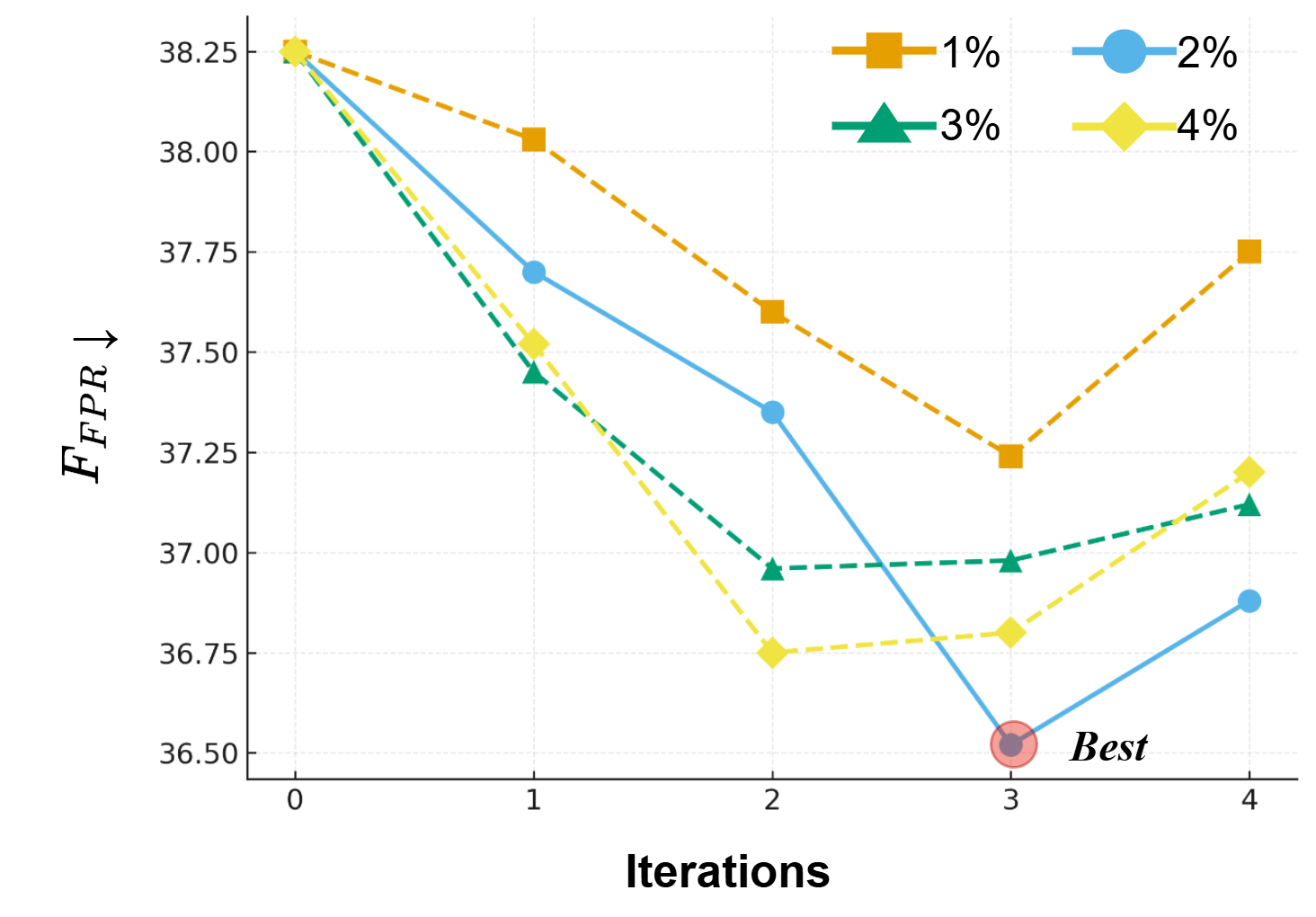}
\vspace{-1mm}
\caption{Analysis of the impact of different decoupling iterations and decoupling ratios on the fairness performance (\(F_{FPR}\)) for ResNet-50 backnone.}
\label{fig:ablation_resnet}
\vspace{-2mm}
\end{figure}

Fig.~\ref{fig:ablation_resnet} presents the analysis of how different decoupling iterations and decoupling ratios within the structural fairness decoupling module affect fairness performance. The results indicate that the optimal trade-off is achieved with a 2\% decoupling ratio at the third iteration, which is consistent with the findings for the Xception backbone.

\section{End-to-end Training Algorithm}
\label{sec:training_algorithm}
The following presents the pseudocode of our training optimization procedure, which integrates Structural Fairness Decoupling and Global Distribution Alignment and implements them throughout the end-to-end training process.

\begin{table}[!htbp]
\centering
\renewcommand{\arraystretch}{1.1}
\begin{tabular}{p{0.96\linewidth}}
\toprule
\textbf{Algorithm 1: Training Optimization} \\
\midrule
\textbf{Input:} Training dataset $\mathcal{D}$ with sensitive attributes, 
pre-trained model, 
max\_iterations, num\_epoch, num\_batch, learning rate $\eta$, 
fairness weight $\lambda_{\text{fair}}$, 
decoupling ratio $pr_c$, 
Sinkhorn regularization coefficient $\varepsilon$, 
and a set of subgroups $\mathcal{J}$. \\[2pt]

\textbf{Output:} A deepfake detection model with improved fairness, parameterized by $\theta_l$. \\[2pt]

\textbf{Initialization:} $\theta_0$, $l = 0$. \\[2pt]

\textbf{for} $e = 1$ to max\_iterations \textbf{do} \\[1pt]
\quad For each channel $k$ in the last convolutional layer, compute its fairness index $F_k$ \\
\quad\quad based on Eq.~\ref{eq:snnl_loss} using $\mathcal{D}$; \\
\quad Select $pr_c$ percent of the channels with the smallest $F_k$ as the decoupling index set $\mathcal{C}^{(e)}$; \\[1pt]
\quad Apply channel decoupling to the last convolutional layer using $\mathcal{C}^{(e)}$; \\[1pt]
\quad \textbf{for} epoch $= 1$ to num\_epoch \textbf{do} \\
\quad\quad \textbf{for} $b = 1$ to num\_batch \textbf{do} \\
\quad\quad\quad Sample a mini-batch $\mathcal{D}_b$ from $\mathcal{D}$; \\
\quad\quad\quad Compute the classification loss $\mathcal{L}_{\mathrm{cls}}$ on $\mathcal{D}_b$ based on Eq.~\ref{eq:classification_loss}; \\
\quad\quad\quad Sample an intersectional subgroup from $\mathcal{J}$ and obtain its predictions; \\
\quad\quad\quad Compute the fairness loss $\mathcal{L}_{\mathrm{fair}}$ based on Eq.~\ref{eq:fairness_loss}; \\
\quad\quad\quad Compute the total loss 
$\mathcal{L} = \mathcal{L}_{\mathrm{cls}} + \lambda_{\text{fair}} \mathcal{L}_{\mathrm{fair}}$; \\
\quad\quad\quad Update parameters 
$\theta_{l+1} \leftarrow \theta_{l} - \eta \nabla_{\theta} \mathcal{L}$; \\
\quad\quad\quad $l \leftarrow l + 1$; \\
\quad\quad \textbf{end} \\
\quad \textbf{end} \\
\textbf{end} \\[2pt]

\textbf{return} $\theta_l$ \\
\bottomrule
\end{tabular}
\end{table}

\section{Experimental Results of Different Training Set}
\label{sec:trained_on_celeb}
To verify that the proposed method also exhibits superior performance when trained on other datasets, we additionally train several methods on the DFDC dataset, using Xception and ResNet-50 as backbone networks, respectively. Since PG-FDD~\cite{lin2024preserving} requires a specific label for training, which is not provided in DFDC dataset, and RSEF-FDD~\cite{he2024redundant} requires additional redundant samples that we are unable to construct from DFDC dataset, these two methods are not included in this comparison. The results for the Xception-based and ResNet-50-based models are reported in Tab.~\ref{tab:dfdc_xception} and Tab.~\ref{tab:dfdc_resnet}, respectively, and demonstrate that, when trained on DFDC dataset, our method still achieves superior fairness across the four test datasets while simultaneously attaining the highest detection performance.

\begin{table*}[!htbp]
\centering
\caption{Evaluation of methods with Xception backbone across intra-domain (DFDC) and cross-domain (FF++, Celeb-DF, DFD) datasets.}
\label{tab:dfdc_xception}
\resizebox{\textwidth}{!}{
\begin{tblr}{
  colspec = {c|c|c|c c c|c c c|c c c|c},
  rowsep = 0pt,
  cells  = {c, valign=m},
  cell{1}{1} = {r=3}{},    
  cell{1}{2} = {r=3}{},    
  cell{1}{3} = {r=3}{},    
  cell{1}{4}  = {c=9}{halign=c, valign=m}, 
  cell{2}{4}  = {c=3}{},   
  cell{2}{7}  = {c=3}{},   
  cell{2}{10} = {c=3}{},   
  cell{4}{1}  = {r=5}{},   
  cell{9}{1}  = {r=5}{},   
  cell{14}{1} = {r=5}{},   
  cell{19}{1} = {r=5}{},   
  row{8,13,18,23} = {bg=mygray},
  hline{1,24} = {-}{0.08em},      
  hline{2,3}  = {1-13}{},         
  hline{4,9,14,19} = {1-13}{},    
}
Datasets & Methods & Backbone & Fairness Metrics(\%) &  &  &  &  &  &  &  &  & {\small Detection Metrics(\%)} \\
         &         &          & Gender &  &  & Race &  &  & Intersection &  &  & {\small Overall} \\
         &         &          & \(F_{FPR}\)↓ & \(F_{DP}\)↓ & \(es-AUC\)↑ & \(F_{FPR}\)↓ & \(F_{DP}\)↓ & \(es-AUC\)↑ & \(F_{FPR}\)↓ & \(F_{DP}\)↓ & \(es-AUC\)↑ & \(AUC\)↑ \\
DFDC     & Ori     & Xception & \underline{1.13} & 7.54 & 85.90 & \underline{13.95} & \underline{23.59} & 70.76 & 68.76 & 40.34 & 52.91 & 87.48 \\
         & DAG-FDD \textsubscript{\textcolor{blue}{WACV'24}}~\cite{ju2024improving} & Xception & 2.27 & 6.69 & 82.27 & 18.43 & 25.47 & 64.28 & 68.90 & 39.55 & 47.44 & 85.31 \\
         & DAW-FDD \textsubscript{\textcolor{blue}{WACV'24}}~\cite{ju2024improving} & Xception & 1.74 & 4.47 & 78.35 & 35.46 & 32.22 & 72.19 & 72.45 & \underline{35.20} & 53.75 & 78.81 \\
         & Fairadapter \textsubscript{\textcolor{blue}{ICASSP'25}}~\cite{ding2025fairadapter} & ViT-L/14 & 3.88 & \underline{3.87} & \underline{87.98} & 23.35 & 39.60 & \underline{80.26} & \underline{49.47} & 43.77 & \underline{72.71} & \underline{88.87} \\
         & Ours    & Xception & \textbf{0.87} & \textbf{3.10} & \textbf{88.50} & \textbf{7.67} & \textbf{22.80} & \textbf{82.59} & \textbf{17.67} & \textbf{27.57} & \textbf{72.90} & \textbf{89.42} \\
FF++     & Ori     & Xception & 16.14 & 16.35 & 55.29 & 22.42 & 16.94 & \textbf{54.16} & 81.51 & 33.20 & \textbf{47.02} & 56.06 \\
         & DAG-FDD \textsubscript{\textcolor{blue}{WACV'24}}~\cite{ju2024improving} & Xception & 18.00 & 23.36 & 56.06 & \underline{17.45} & \underline{11.17} & 50.60 & \underline{69.19} & \underline{29.73} & \underline{46.13} & 56.82 \\
         & DAW-FDD \textsubscript{\textcolor{blue}{WACV'24}}~\cite{ju2024improving} & Xception & 9.67 & 6.94 & 52.78 & 29.98 & 13.43 & 46.21 & 99.66 & 30.44 & 36.95 & 54.78 \\
         & Fairadapter \textsubscript{\textcolor{blue}{ICASSP'25}}~\cite{ding2025fairadapter} & ViT-L/14 & \underline{7.19} & \underline{4.74} & \underline{58.84} & 33.80 & 16.94 & 51.05 & 73.72 & 29.96 & 45.53 & \underline{59.64} \\
         & Ours    & Xception & \textbf{1.01} & \textbf{1.33} & \textbf{59.82} & \textbf{9.00} &\textbf{ 6.26} & \underline{51.34} & \textbf{19.60} & \textbf{8.34} & 42.82 & \textbf{60.51} \\
Celeb-DF & Ori     & Xception & 9.13 & 15.43 & \underline{62.33} & \underline{11.32} & \underline{16.95} & \underline{57.44} & \underline{14.71} & \underline{16.19} & \underline{57.27} & \underline{66.09} \\
         & DAG-FDD \textsubscript{\textcolor{blue}{WACV'24}}~\cite{ju2024improving} & Xception & \textbf{1.50} & \underline{7.26} & 51.87 & 17.23 & 18.58 & 46.63 & 20.68 & 18.02 & 44.94 & 56.43 \\
         & DAW-FDD \textsubscript{\textcolor{blue}{WACV'24}}~\cite{ju2024improving} & Xception & 19.77 & 25.13 & 53.36 & 24.58 & 23.85 & 42.21 & 23.83 & 25.55 & 42.93 & 58.82 \\
         & Fairadapter \textsubscript{\textcolor{blue}{ICASSP'25}}~\cite{ding2025fairadapter} & ViT-L/14 & 13.29 & 13.72 & 49.79 & 29.12 & 17.30 & 46.76 & 15.99 & 17.46 & 40.75 & 54.01 \\
         & Ours    & Xception & \underline{3.07} & \textbf{6.96} & \textbf{62.51} & \textbf{10.32} & \textbf{15.94} & \textbf{59.56} & \textbf{8.74} & \textbf{13.44} & \textbf{58.36} & \textbf{70.21} \\
DFD      & Ori     & Xception & 7.81 & 11.44 & 61.91 & 13.28 & \underline{6.19} & 62.11 & 48.35 & \underline{17.02} & 45.47 & 67.51 \\
         & DAG-FDD \textsubscript{\textcolor{blue}{WACV'24}}~\cite{ju2024improving} & Xception & \textbf{0.36} & \underline{11.35} & 65.12 & \underline{13.02} & 10.39 & 63.57 & \underline{27.30} & 24.24 & 57.27 & 68.20 \\
         & DAW-FDD \textsubscript{\textcolor{blue}{WACV'24}}~\cite{ju2024improving} & Xception & 14.89 & 13.39 & 50.10 & 14.67 & 15.89 & 47.04 & 45.36 & 37.80 & 44.58 & 53.71 \\
         & Fairadapter \textsubscript{\textcolor{blue}{ICASSP'25}}~\cite{ding2025fairadapter} & ViT-L/14 & 15.12 & 12.40 & \underline{68.17} & 28.18 & 9.26 & \underline{65.72} & 48.29 & 38.00 & \underline{57.65} & \underline{70.78} \\
         & Ours    & Xception & \underline{6.25} & \textbf{10.94} & \textbf{68.61} & \textbf{12.33} & \textbf{6.09} & \textbf{66.87} & \textbf{19.26} & \textbf{12.00} & \textbf{60.95} & \textbf{70.83} \\
\end{tblr}
}
\end{table*}
\begin{table*}[!htbp]
\centering
\caption{Evaluation of methods with ResNet-50 backbone across intra-domain (DFDC) and cross-domain (FF++, Celeb-DF, DFD) datasets.}
\label{tab:dfdc_resnet}
\resizebox{\textwidth}{!}{
\begin{tblr}{
  colspec = {c|c|c|c c c|c c c|c c c|c},
  rowsep = 0pt,
  cells  = {c, valign=m},
  cell{1}{1} = {r=3}{},    
  cell{1}{2} = {r=3}{},    
  cell{1}{3} = {r=3}{},    
  cell{1}{4}  = {c=9}{halign=c, valign=m}, 
  cell{2}{4}  = {c=3}{},   
  cell{2}{7}  = {c=3}{},   
  cell{2}{10} = {c=3}{},   
  cell{4}{1}  = {r=5}{},   
  cell{9}{1}  = {r=5}{},   
  cell{14}{1} = {r=5}{},   
  cell{19}{1} = {r=5}{},   
  row{8,13,18,23} = {bg=mygray},
  hline{1,24} = {-}{0.08em},      
  hline{2,3}  = {1-13}{},         
  hline{4,9,14,19} = {1-13}{},    
}
Datasets & Methods & Backbone & Fairness Metrics(\%) &  &  &  &  &  &  &  &  & {\small Detection Metrics(\%)} \\
         &         &          & Gender &  &  & Race &  &  & Intersection &  &  & {\small Overall} \\
         &         &          & \(F_{FPR}\)↓ & \(F_{DP}\)↓ & \(es-AUC\)↑ & \(F_{FPR}\)↓ & \(F_{DP}\)↓ & \(es-AUC\)↑ & \(F_{FPR}\)↓ & \(F_{DP}\)↓ & \(es-AUC\)↑ & \(AUC\)↑ \\
DFDC     & Ori     & ResNet-50 & 3.19 & 12.60 & 91.66 & 10.18 & 28.48 & \underline{83.38} & 39.22 & 40.33 & 69.57 & 92.43 \\
         & DAG-FDD \textsubscript{\textcolor{blue}{WACV'24}}~\cite{ju2024improving} & ResNet-50 & 2.30 & 14.24 & \underline{91.76} & \underline{9.06} & \textbf{27.39} & 81.53 & \underline{27.43} & \underline{38.95} & 66.70 & \underline{92.47} \\
         & DAW-FDD \textsubscript{\textcolor{blue}{WACV'24}}~\cite{ju2024improving} & ResNet-50 & \underline{2.06} & 12.89 & 90.56 & 11.13 & 28.46 & 80.34 & 29.95 & 39.83 & 64.24 & 92.01 \\
         & Fairadapter \textsubscript{\textcolor{blue}{ICASSP'25}}~\cite{ding2025fairadapter} & ViT-L/14 & 3.88 & \underline{10.87} & 87.98 & 23.35 & 39.60 & 80.26 & 49.47 & 43.77 & \textbf{72.71} & 88.87 \\
         & Ours    & ResNet-50 & \textbf{1.44} & \textbf{10.44} & \textbf{92.54} & \textbf{7.62} & \underline{28.23} & \textbf{84.99} & \textbf{20.69} & \textbf{37.19} & \underline{72.12} & \textbf{93.92} \\
FF++     & Ori     & ResNet-50 & 10.77 & 11.89 & 57.26 & 41.89 & 22.92 & 50.34 & 95.46 & 35.10 & 42.76 & 57.66 \\
         & DAG-FDD \textsubscript{\textcolor{blue}{WACV'24}}~\cite{ju2024improving} & ResNet-50 & 8.68 & \underline{9.44} & 58.72 & \underline{21.09} & 19.06 & 50.24 & \underline{57.90} & \underline{22.92} & 43.89 & 59.22 \\
         & DAW-FDD \textsubscript{\textcolor{blue}{WACV'24}}~\cite{ju2024improving} & ResNet-50 & 14.85 & 16.54 & 57.38 & 21.41 & \underline{12.57} & 48.82 & 62.30 & 32.43 & 42.63 & 57.61 \\
         & Fairadapter \textsubscript{\textcolor{blue}{ICASSP'25}}~\cite{ding2025fairadapter} & ViT-L/14 & \underline{8.59} & \textbf{4.74} & \underline{58.84} & 33.80 & 16.94 & \underline{51.05} & 73.72 & 29.96 & \underline{45.53} & \underline{59.64} \\
         & Ours    & ResNet-50 & \textbf{8.49} & 13.02 & \textbf{59.89} & \textbf{10.52} & \textbf{6.05} & \textbf{54.18} & \textbf{39.70} & \textbf{22.29} & \textbf{45.75} & \textbf{60.25} \\
Celeb-DF & Ori     & ResNet-50 & 3.14 & 5.40 & 59.20 & \underline{26.33} & 29.42 & \textbf{62.44} & \underline{27.29} & \underline{19.14} & 50.72 & \underline{66.15} \\
         & DAG-FDD \textsubscript{\textcolor{blue}{WACV'24}}~\cite{ju2024improving} & ResNet-50 & \textbf{0.21} & 5.27 & \underline{60.49} & 26.52 & \underline{27.08} & 55.19 & 31.88 & 29.02 & \underline{53.40} & 63.78 \\
         & DAW-FDD \textsubscript{\textcolor{blue}{WACV'24}}~\cite{ju2024improving} & ResNet-50 & 3.79 & \underline{4.41} & 58.81 & 33.51 & 37.95 & 49.78 & 35.01 & 35.73 & 48.55 & 64.02 \\
         & Fairadapter \textsubscript{\textcolor{blue}{ICASSP'25}}~\cite{ding2025fairadapter} & ViT-L/14 & 13.29 & 13.72 & 49.79 & 29.12 & \textbf{17.30} & 46.76 & 28.99 & \textbf{17.46} & 40.75 & 54.01 \\
         & Ours    & ResNet-50 & \underline{2.32} & \textbf{4.38} & \textbf{65.82} & \textbf{25.96} & 39.51 & \underline{56.70} & \textbf{26.70} & 34.79 & \textbf{55.72} & \textbf{73.28} \\
DFD      & Ori     & ResNet-50 & 11.89 & 7.88 & \underline{66.20} & 11.07 & 24.81 & 53.77 & 73.70 & 43.89 & 46.76 & 63.76 \\
         & DAG-FDD \textsubscript{\textcolor{blue}{WACV'24}}~\cite{ju2024improving} & ResNet-50 & \underline{5.22} & \underline{6.85} & 57.48 & 6.69 & \underline{17.95} & 50.66 & \underline{34.71} & \underline{26.31} & 46.82 & 58.75 \\
         & DAW-FDD \textsubscript{\textcolor{blue}{WACV'24}}~\cite{ju2024improving} & ResNet-50 & 18.99 & 18.81 & 60.99 & \underline{4.02} & 22.74 & 52.84 & 50.34 & 45.11 & \underline{49.43} & \underline{64.03} \\
         & Fairadapter \textsubscript{\textcolor{blue}{ICASSP'25}}~\cite{ding2025fairadapter} & ViT-L/14 & 15.12 & 12.40 & \textbf{68.17} & 28.18 & 19.26 & \underline{54.72} & 48.29 & 38.00 & \textbf{57.65} & 60.78 \\
         & Ours    & ResNet-50 & \textbf{0.61} & \textbf{0.88} & 63.24 & \textbf{1.08} & \textbf{13.29} & \textbf{54.91} & \textbf{12.20} & \textbf{15.93} & 45.08 & \textbf{64.35} \\
\end{tblr}
}
\end{table*}

\section{Limitation and Future Work}
\label{sec:limitation}
One limitation of our method lies in its reliance on datasets with accurate demographic annotations. Currently, fairness-related annotations in existing datasets are quite limited; therefore, our fairness analysis can only focus on gender and race. In future work, we plan to enrich existing datasets with more fairness-related labels, and even construct our own dataset to enable more comprehensive and in-depth research.

\section{Effect of Trade-off Hyperparameter $\lambda$}
\label{sec:effect_lambda}
\begin{figure}[!htbp]
\centering
\includegraphics[width=0.8\linewidth]{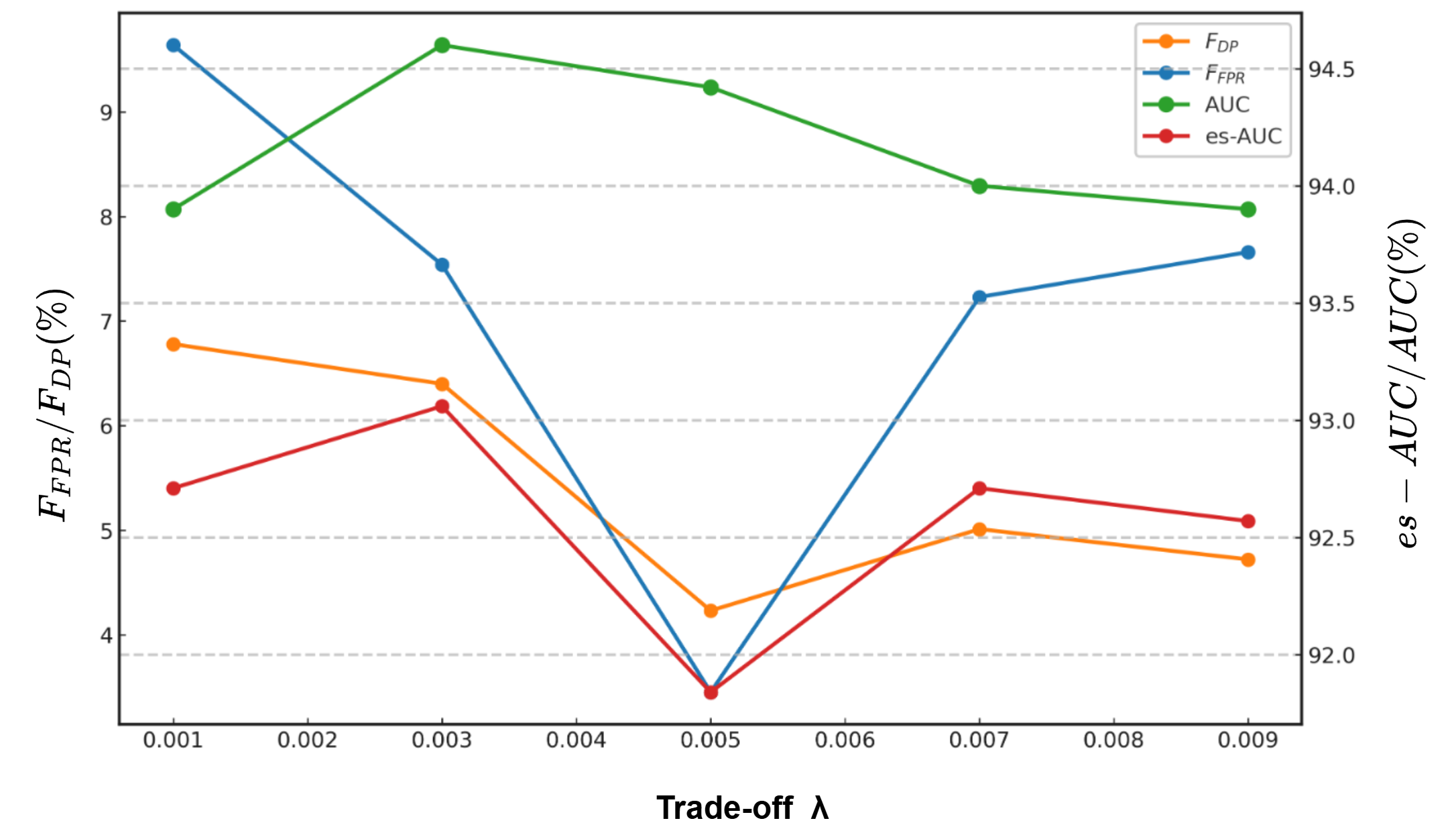}
\vspace{-1mm}
\caption{Sensitivity analysis of parameter $\lambda$ on the trade-off between fairness and detection accuracy for the gender attribute on FF++.}
\label{fig:optimal_lambda}
\vspace{-1mm}
\end{figure}

\begin{figure}[!htbp]
\centering
\includegraphics[width=\linewidth]{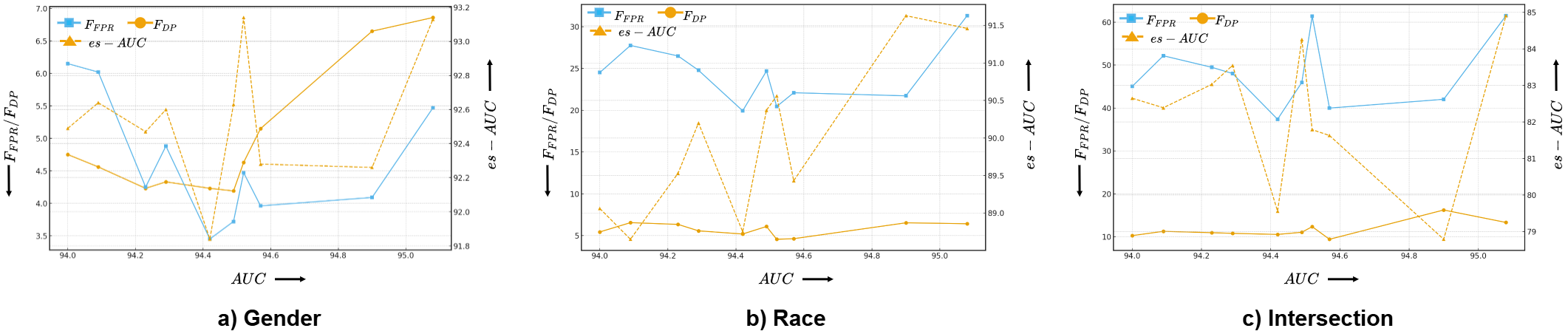}
\vspace{-1mm}
\caption{Trends in Fairness Metrics vs. AUC Score. From left to right, the chart shows how fairness metrics for gender, race, and intersection attribute change with AUC, illustrating the trade-off between accuracy and fairness.}
\label{fig:metric_auc}
\vspace{-1mm}
\end{figure}

To verify the effect of the trade-off hyperparameter in Eq.~\ref{eq:overall loss}, we conducted sensitivity analysis on the FF++ dataset. Fig.~\ref{fig:optimal_lambda} shows the fairness metrics and detection metric AUC for different $\lambda$ values. The experimental results indicate that when $\lambda$ is set to 0.005, the model achieves optimal fairness performance while maintaining fair AUC scores. It is noteworthy that the analysis reveals a trade-off between fairness and AUC scores: as $\lambda$ increases from 0.003 to 0.005, AUC decreases, but fairness improves. To more clearly illustrate the relationship between each fairness metric and AUC, we present these dynamic changes separately in Fig.~\ref{fig:metric_auc}, which displays the trend of increased AUC corresponding to reduced fairness.

\section{The T-SNE Visualization of Demographic Features}
\label{sec:tsne_features}
\begin{figure}[!htbp]
\centering
\includegraphics[width=0.8\linewidth]{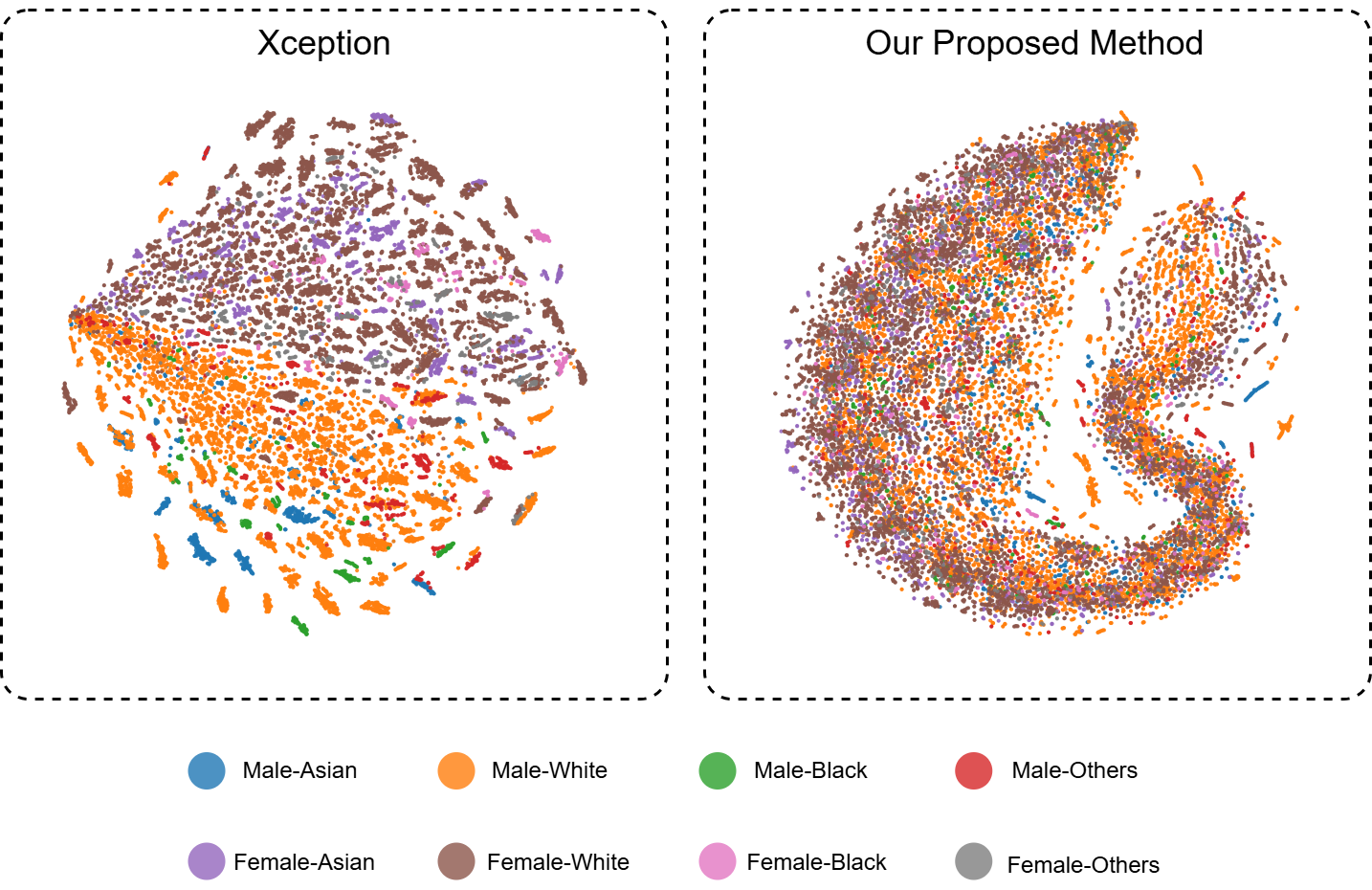}
\vspace{-1mm}
\caption{The T-SNE visualization of demographic features extracted from Xception and our method on FF++ dataset.}
\label{fig:T-SNE}
\vspace{-1mm}
\end{figure}

We present in Fig.~\ref{fig:T-SNE} the t-SNE visualizations of demographic features extracted on FF++ by the vanilla Xception model and by our method, respectively. In the visualization, the different intersectional demographic groups extracted by the unconstrained Xception model form clearly separated clusters, whereas the features extracted by our method show these intersectional groups intermingled in the feature space. This indicates that our model has discovered common fingerprints across demographics, leading to a more fair representation. The t-SNE results further reveal that most subgroups in FF++ belong to the Male-White and Female-White categories, highlighting a strong dataset bias that makes fair detection particularly challenging and underscoring the necessity of our dual-mechanism synergistic optimization framework to improve fairness for minority subgroups.

\section{Additional Experimental Results}
\label{sec:more_experiments}
For a fairer comparison, we conducted additional experiments on a recent public fair forgery detection benchmark~\cite{liu2025thinking} and included more comparison methods under our experimental settings. The results in Tables~\ref{tab:fairfd1} and~\ref{tab:fairfd2} show that our method still outperforms all compared methods on this benchmark. The results on all four datasets, with more comparison methods included, consistently show that our method outperforms the other compared methods.

\begin{table*}[!htbp]
\centering
\scriptsize
\setlength{\tabcolsep}{2.6pt}
\renewcommand{\arraystretch}{1.15}
\caption{Additional experimental results on the FairFD benchmark.
 Best results are in \textbf{bold} and second best are \underline{underlined}.}
\resizebox{\textwidth}{!}{
\begin{tabular}{llcccccccccccccc}
\toprule
\multicolumn{2}{c}{\multirow{2}{*}{Fairness Metric}} &
\multicolumn{6}{c}{Spatial-based} &
\multicolumn{3}{c}{Frequency-based} &
\multicolumn{5}{c}{Fairness-enhanced} \\
\cmidrule(lr){3-8}\cmidrule(lr){9-11}\cmidrule(lr){12-16}
\multicolumn{2}{c}{} &
Xception & RECCE & UCF & Capsule & FFD & CORE &
F3Net & SPSL & SRM &
DAG & DAW & PFGDFD & RSEF-FDD & Ours \\
\midrule
\multirow{4}{*}{Naive Metric} & DPD$\downarrow$     & 0.1810 & 0.1338 & 0.1765 & 0.0969 & 0.1099 & 0.0951 & 0.0674 & \underline{0.0203} & 0.0990 & 0.1723 & 0.0513 & 0.0805 & 0.0378 & \textbf{0.0159} \\
 & DEOdds$\downarrow$  & 0.1666 & 0.1264 & 0.1495 & 0.0902 & 0.1005 & 0.0798 & 0.0763 & \underline{0.0304} & 0.0714 & 0.2288 & 0.0593 & 0.1396 & 0.0693 & \textbf{0.0118} \\
 & DEO$\downarrow$     & 0.2088 & 0.1548 & 0.2014 & 0.1118 & 0.1242 & 0.1084 & 0.0801 & \underline{0.0215} & 0.1090 & 0.2105 & 0.0611 & 0.1032 & 0.0513 & \textbf{0.0129} \\
 & STD$\downarrow$     & 0.0647 & 0.0474 & 0.0631 & 0.0343 & 0.0398 & 0.0342 & 0.0265 & \textbf{0.0080} & 0.0355 & 0.0636 & 0.0195 & 0.0328 & \underline{0.0134} & 0.0148 \\
\midrule
\multirow{4}{*}{Approach Averaged} & AADPD$\downarrow$    & 0.2024 & 0.1572 & 0.2175 & 0.1323 & 0.1552 & 0.1147 & 0.1158 & \underline{0.0556} & 0.1413 & 0.2201 & 0.0735 & 0.1393 & 0.0707 & \textbf{0.0438} \\
 & AADEOdds$\downarrow$ & 0.1669 & 0.1302 & 0.1630 & 0.1034 & 0.1196 & 0.0858 & 0.0961 & \underline{0.0481} & 0.0925 & 0.2324 & 0.0662 & 0.1560 & 0.0755 & \textbf{0.0306} \\
 & AADEO$\downarrow$     & 0.2095 & 0.1626 & 0.2284 & 0.1381 & 0.1623 & 0.1205 & 0.1197 & \underline{0.0571} & 0.1511 & 0.2177 & 0.0749 & 0.1360 & 0.0738 & \textbf{0.0433} \\
 & AASTD$\downarrow$    & 0.0750 & 0.0578 & 0.0809 & 0.0493 & 0.0576 & 0.0449 & 0.0448 & \underline{0.0219} & 0.0531 & 0.0834 & 0.0283 & 0.0530 & 0.0273 & \textbf{0.0154} \\
\midrule
\multirow{4}{*}{Utility Regularized} & URDPD$\downarrow$    & 0.1357 & 0.1118 & 0.1523 & 0.0808 & 0.1037 & 0.0803 & 0.0806 & \underline{0.0320} & 0.0904 & 0.1474 & 0.0555 & 0.0881 & 0.0562 & \textbf{0.0255} \\
 & URDEOdds$\downarrow$ & 0.1057 & 0.0852 & 0.1069 & 0.0639 & 0.0763 & 0.0567 & 0.0625 & \underline{0.0299} & 0.0584 & 0.1445 & 0.0440 & 0.0986 & 0.0495 & \textbf{0.0210} \\
 & URDEO$\downarrow$     & 0.1417 & 0.1171 & 0.1614 & 0.0842 & 0.1092 & 0.0850 & 0.0842 & \underline{0.0324} & 0.0968 & 0.1480 & 0.0578 & 0.0860 & 0.0595 & \textbf{0.0264} \\
 & URSTD$\downarrow$    & 0.0501 & 0.0410 & 0.0565 & 0.0301 & 0.0384 & 0.0313 & 0.0312 & \underline{0.0126} & 0.0339 & 0.0559 & 0.0214 & 0.0335 & 0.0216 & \textbf{0.0119} \\
\midrule
Utility & AUC$\uparrow$  & 0.6911 & 0.6897 & \underline{0.7214} & 0.6815 & \textbf{0.7304} & 0.6864 & 0.6564 & 0.6763 & 0.7102 & 0.6672 & 0.6604 & 0.6302 & 0.6413 & 0.7026 \\
\bottomrule
\end{tabular}
}
\end{table*}
\label{tab:fairfd1}

\begin{table*}[!htbp]
\centering
\scriptsize
\setlength{\tabcolsep}{2.2pt}
\renewcommand{\arraystretch}{1.15}
\caption{Additional experimental results on the FairFD benchmark, where “-BPFA” denotes models processed using the benchmark’s BPFA method. Best results are in \textbf{bold} and second best are \underline{underlined}.}
\resizebox{\textwidth}{!}{
\begin{tabular}{llcccccccccccccc}
\toprule
\multicolumn{2}{c}{\multirow{2}{*}{Fairness Metric}} &
\multicolumn{6}{c}{Spatial-based} &
\multicolumn{3}{c}{Frequency-based} &
\multicolumn{5}{c}{Fairness-enhanced} \\
\cmidrule(lr){3-8}\cmidrule(lr){9-11}\cmidrule(lr){12-16}
\multicolumn{2}{c}{} &
Xception-BPFA & RECCE-BPFA & UCF-BPFA & Capsule-BPFA & FFD-BPFA & CORE-BPFA &
F3Net-BPFA & SPSL-BPFA & SRM-BPFA &
DAG-BPFA & DAW-BPFA & PFGDFD-BPFA & RSEF-FDD & Ours \\
\midrule

\multirow{4}{*}{Naive Metric} & DPD$\downarrow$ & 0.1644 & 0.1328 & 0.1705 & 0.0951 & 0.1096 & 0.0946 & 0.0674 & \underline{0.0181} & 0.0993 & 0.1286 & 0.0510 & 0.0594 & 0.0378 & \textbf{0.0159} \\
 & DEOdds$\downarrow$ & 0.1532 & 0.1327 & 0.1571 & 0.1011 & 0.0999 & 0.0837 & 0.0762 & \underline{0.0209} & 0.0732 & 0.1807 & 0.0553 & 0.1337 & 0.0693 & \textbf{0.0118} \\
 & DEO$\downarrow$ & 0.1899 & 0.1550 & 0.1967 & 0.1119 & 0.1237 & 0.1085 & 0.0801 & \underline{0.0200} & 0.1090 & 0.1587 & 0.0602 & 0.0796 & 0.0513 & \textbf{0.0129} \\
 & STD$\downarrow$ & 0.0590 & 0.0471 & 0.0608 & 0.0337 & 0.0397 & 0.0341 & 0.0264 & \textbf{0.0072} & 0.0356 & 0.0457 & 0.0198 & 0.0238 & \underline{0.0134} & 0.0148 \\
\midrule

\multirow{4}{*}{Approach Averaged} & AADPD$\downarrow$ & 0.1854 & 0.1584 & 0.2159 & 0.1342 & 0.1546 & 0.1155 & 0.1155 & \underline{0.0473} & 0.1417 & 0.1648 & 0.0774 & 0.1079 & 0.0707 & \textbf{0.0438} \\
 & AADEOdds$\downarrow$ & 0.1540 & 0.1366 & 0.1711 & 0.1143 & 0.1189 & 0.0898 & 0.0959 & \underline{0.0357} & 0.0943 & 0.1820 & 0.0651 & 0.1442 & 0.0755 & \textbf{0.0306} \\
 & AADEO$\downarrow$ & 0.1917 & 0.1628 & 0.2249 & 0.1382 & 0.1617 & 0.1206 & 0.1195 & \underline{0.0496} & 0.1511 & 0.1614 & 0.0799 & 0.1006 & 0.0738 & \textbf{0.0433} \\
 & AASTD$\downarrow$ & 0.0693 & 0.0582 & 0.0803 & 0.0501 & 0.0574 & 0.0450 & 0.0447 & \underline{0.0182} & 0.0532 & 0.0613 & 0.0298 & 0.0411 & 0.0273 & \textbf{0.0154} \\
\midrule

\multirow{4}{*}{Utility Regularized} & URDPD$\downarrow$ & 0.1218 & 0.1125 & 0.1507 & 0.0820 & 0.1032 & 0.0807 & 0.0804 & \underline{0.0265} & 0.0906 & 0.1107 & 0.0565 & 0.0644 & 0.0562 & \textbf{0.0255} \\
 & URDEOdds$\downarrow$ & 0.0964 & 0.0888 & 0.1109 & 0.0708 & 0.0758 & 0.0589 & 0.0623 & \underline{0.0218} & 0.0595 & 0.1131 & 0.0430 & 0.0969 & 0.0495 & \textbf{0.0210} \\
 & URDEO$\downarrow$ & 0.1268 & 0.1173 & 0.1587 & 0.0843 & 0.1087 & 0.0851 & 0.0840 & \underline{0.0275} & 0.0968 & 0.1102 & 0.0592 & 0.0578 & 0.0595 & \textbf{0.0264} \\
 & URSTD$\downarrow$ & 0.0454 & 0.0413 & 0.0559 & 0.0306 & 0.0382 & 0.0313 & 0.0311 & \textbf{0.0102} & 0.0339 & 0.0412 & 0.0218 & 0.0245 & 0.0216 & \underline{0.0119} \\
\midrule

Utility & AUC$\uparrow$ & 0.6952 & 0.6877 & \underline{0.7226} & 0.6794 & \textbf{0.7305} & 0.6874 & 0.6577 & 0.6862 & 0.7100 & 0.6512 & 0.6668 & 0.6445 & 0.6413 & 0.7026 \\
\bottomrule
\end{tabular}
}
\end{table*}
\label{tab:fairfd2}


\begin{table*}[t]
\centering
\scriptsize
\setlength{\tabcolsep}{2.2pt}
\renewcommand{\arraystretch}{1.15}
\caption{Additional experimental results on the FF++ dataset with more included methods. Best results are in \textbf{bold} and second best are \underline{underlined}.}
\resizebox{\textwidth}{!}{
\begin{tabular}{llccccccccccccccc}
\toprule
\multicolumn{2}{c}{\multirow{2}{*}{Fairness Metric}} &
\multicolumn{6}{c}{Spatial-based} &
\multicolumn{3}{c}{Frequency-based} &
\multicolumn{6}{c}{Fairness-enhanced} \\
\cmidrule(lr){3-8}\cmidrule(lr){9-11}\cmidrule(lr){12-17}
\multicolumn{2}{c}{} &
Xception & RECCE & UCF & Capsule & FFD & CORE &
F3Net & SPSL & SRM &
DAG & DAW & PFGDFD & Fairadapter & RSEF-FDD & Ours \\
\midrule

\multirow{3}{*}{Gender} & $F_{FPR}\downarrow$ & 4.10 & 0.64 & 8.87 & 34.54 & 3.07 & 1.51 & 2.05 & 11.51 & 10.06 & 1.82 & 0.78 & 0.62 & 4.16 & \underline{0.57} & \textbf{0.53} \\
 & $F_{DP}\downarrow$ & 5.72 & 3.88 & 6.44 & 18.65 & 3.88 & 3.89 & 4.60 & \underline{3.72} & 4.24 & 4.65 & 9.52 & 4.74 & 12.21 & 8.55 & \textbf{3.61} \\
 & $es-AUC\uparrow$ & 91.93 & 85.76 & 86.38 & 74.85 & 85.84 & 85.88 & 88.23 & 87.21 & 84.99 & 94.87 & 95.76 & \underline{96.32} & 67.85 & 94.91 & \textbf{96.45} \\
\midrule

\multirow{3}{*}{Race} & $F_{FPR}\downarrow$ & 19.76 & 26.05 & 32.02 & 44.47 & 23.22 & 10.81 & 16.20 & 35.03 & 26.13 & \underline{5.48} & \textbf{5.43} & 11.13 & 43.22 & 8.39 & 9.29 \\
 & $F_{DP}\downarrow$ & 4.74 & 4.66 & 4.55 & 8.34 & 4.86 & \underline{4.42} & 5.33 & 5.13 & 4.79 & 9.20 & 14.69 & 4.78 & 20.39 & 5.28 & \textbf{4.35} \\
 & $es-AUC\uparrow$ & 82.85 & 80.63 & 84.53 & 64.57 & 80.82 & 82.62 & 83.75 & 81.06 & 81.98 & 93.43 & 94.15 & \underline{94.52} & 56.03 & 93.60 & \textbf{94.86} \\
\midrule

\multirow{3}{*}{Intersection} & $F_{FPR}\downarrow$ & 36.03 & 61.92 & 73.40 & 141.99 & 50.39 & 26.62 & 35.04 & 37.73 & 82.75 & 24.08 & \underline{14.36} & \textbf{9.19} & 86.91 & 23.64 & 20.18 \\
 & $F_{DP}\downarrow$ & 14.64 & 10.05 & 10.07 & 26.62 & 10.78 & 9.98 & 10.07 & 10.35 & \textbf{6.68} & 26.25 & 26.13 & 13.39 & 42.44 & 21.74 & \underline{9.47} \\
 & $es-AUC\uparrow$ & 74.43 & 69.75 & 70.96 & 52.46 & 72.29 & 70.86 & 75.19 & 68.91 & 68.15 & 85.80 & 86.74 & \underline{86.83} & 45.98 & 85.77 & \textbf{86.91} \\
\midrule

Overall & $AUC\uparrow$ & 92.69 & 88.50 & 89.33 & 79.15 & 87.07 & 88.31 & 90.45 & 89.87 & 88.46 & 96.72 & 97.46 & \underline{97.66} & 71.50 & 97.09 & \textbf{97.71} \\
\bottomrule
\end{tabular}
}
\end{table*}

\label{tab:supp_ff++}


\begin{table*}[t]
\centering
\scriptsize
\setlength{\tabcolsep}{2.2pt}
\renewcommand{\arraystretch}{1.15}
\caption{Additional experimental results on the DFD dataset with more included methods. Best results are in \textbf{bold} and second best are \underline{underlined}.}
\resizebox{\textwidth}{!}{%
\begin{tabular}{llccccccccccccccc}
\toprule
\multicolumn{2}{c}{\multirow{2}{*}{Fairness Metric}} &
\multicolumn{6}{c}{Spatial-based} &
\multicolumn{3}{c}{Frequency-based} &
\multicolumn{6}{c}{Fairness-enhanced} \\
\cmidrule(lr){3-8}\cmidrule(lr){9-11}\cmidrule(lr){12-17}
\multicolumn{2}{c}{} &
Xception & RECCE & UCF & Capsule & FFD & CORE &
F3Net & SPSL & SRM &
DAG & DAW & PFGDFD & Fairadapter & RSEF-FDD & Ours \\
\midrule
\multirow{3}{*}{Gender} & $F_{FPR}\downarrow$ & 9.45 & 5.89 & 6.06 & 12.86 & 10.92 & 5.71 & 5.47 & 5.76 & 16.14 & 5.41 & 5.30 & \underline{5.05} & 6.32 & 18.50 & \textbf{4.72} \\
 & $F_{DP}\downarrow$ & 6.63 & 6.40 & 6.02 & 5.69 & 7.76 & 6.91 & 6.63 & 6.84 & 9.23 & 5.60 & 10.63 & \textbf{1.86} & 7.55 & 6.11 & \underline{5.49} \\
 & $es-AUC\uparrow$ & 72.68 & 75.41 & 75.47 & 75.03 & 72.76 & 72.34 & 72.17 & 75.26 & 73.26 & 75.87 & 71.68 & \underline{76.37} & 63.28 & 75.97 & \textbf{78.98} \\
\midrule
\multirow{3}{*}{Race} & $F_{FPR}\downarrow$ & 7.75 & 4.21 & 2.78 & 4.23 & 2.79 & 9.36 & 5.59 & 18.43 & 14.20 & \underline{2.56} & 4.32 & 5.79 & 11.29 & 19.45 & \textbf{2.09} \\
 & $F_{DP}\downarrow$ & 22.31 & 5.62 & 7.14 & \textbf{0.09} & 6.87 & 5.02 & 11.14 & 5.81 & \underline{2.70} & 21.34 & 20.16 & 19.31 & 12.28 & 19.40 & 18.66 \\
 & $es-AUC\uparrow$ & 69.07 & 73.18 & 74.11 & 71.82 & 72.03 & \underline{76.83} & 74.45 & 71.91 & 75.06 & 70.14 & 63.24 & 74.81 & 56.28 & 72.65 & \textbf{77.44} \\
\midrule
\multirow{3}{*}{Intersection} & $F_{FPR}\downarrow$ & 35.06 & 28.46 & 28.50 & 29.00 & 29.75 & 28.33 & \underline{27.73} & 65.35 & 36.50 & 33.00 & 33.81 & 28.00 & 29.56 & 35.34 & \textbf{27.32} \\
 & $F_{DP}\downarrow$ & 25.53 & 23.33 & 28.28 & 28.37 & 26.32 & 22.83 & 28.27 & 23.33 & 25.66 & 23.02 & 24.72 & 23.93 & 33.11 & \textbf{10.97} & \underline{22.81} \\
 & $es-AUC\uparrow$ & 59.20 & 67.25 & 68.89 & 60.86 & 66.01 & 62.22 & 62.82 & 67.12 & \underline{69.37} & 61.09 & 56.01 & 66.32 & 48.63 & 67.24 & \textbf{69.73} \\
\midrule
Overall & $AUC\uparrow$ & 74.32 & 79.86 & 80.93 & 76.48 & 75.26 & 80.97 & 80.77 & 80.51 & \underline{81.22} & 76.29 & 73.71 & 80.70 & 68.12 & 80.54 & \textbf{81.46} \\
\bottomrule
\end{tabular}%
}
\end{table*}
\label{tab:supp_dfd}


\begin{table*}[t]
\centering
\scriptsize
\setlength{\tabcolsep}{2.2pt}
\renewcommand{\arraystretch}{1.15}
\caption{Additional experimental results on the DFDC dataset with more included methods. Best results are in \textbf{bold} and second best are \underline{underlined}.}
\resizebox{\textwidth}{!}{%
\begin{tabular}{llccccccccccccccc}
\toprule
\multicolumn{2}{c}{\multirow{2}{*}{Fairness Metric}} &
\multicolumn{6}{c}{Spatial-based} &
\multicolumn{3}{c}{Frequency-based} &
\multicolumn{6}{c}{Fairness-enhanced} \\
\cmidrule(lr){3-8}\cmidrule(lr){9-11}\cmidrule(lr){12-17}
\multicolumn{2}{c}{} &
Xception & RECCE & UCF & Capsule & FFD & CORE &
F3Net & SPSL & SRM &
DAG & DAW & PFGDFD & Fairadapter & RSEF-FDD & Ours \\
\midrule

\multirow{3}{*}{Gender} & $F_{FPR}\downarrow$  & 8.67 & 6.70 & 7.94 & 7.56 & 6.34 & 5.76 & 4.60 & 3.04 & 10.20 & 5.29 & 3.60 & 2.35 & 2.39 & \underline{2.09} & \textbf{1.76} \\
 & $F_{DP}\downarrow$   & 6.70 & 5.81 & 9.68 & 5.21 & 5.41 & 6.00 & 4.19 & 4.38 & 7.73 & 6.68 & 3.67 & \textbf{2.57} & 4.57 & 3.83 & \underline{3.67} \\
 & $es-AUC\uparrow$  & 54.56 & 56.98 & 55.15 & 56.98 & 50.58 & 51.22 & 57.64 & 57.54 & 51.61 & 57.41 & 55.88 & 57.71 & 55.88 & \underline{57.90} & \textbf{57.94} \\
\midrule

\multirow{3}{*}{Race} & $F_{FPR}\downarrow$  & 19.41 & 19.02 & 14.64 & \underline{6.97} & 12.22 & 11.58 & 16.64 & 20.21 & \textbf{5.76} & 9.26 & 22.36 & 10.10 & 32.49 & 12.24 & 17.93 \\
 & $F_{DP}\downarrow$   & 7.99 & 8.36 & 8.67 & 7.77 & 11.80 & \underline{7.69} & 14.42 & 15.93 & 11.06 & 12.51 & 9.72 & 11.49 & 26.77 & 7.74 & \textbf{7.58} \\
 & $es-AUC\uparrow$  & 47.02 & 46.05 & 51.94 & 42.83 & 49.25 & 51.23 & 45.36 & 46.07 & 49.77 & 45.91 & 47.77 & 50.53 & \underline{52.27} & 50.27 & \textbf{52.33} \\
\midrule

\multirow{3}{*}{Intersection} & $F_{FPR}\downarrow$ & 53.42 & 60.30 & 53.98 & 58.34 & 49.31 & 44.75 & 41.88 & 42.98 & 49.97 & 45.30 & 46.80 & \textbf{32.16} & 74.10 & 38.89 & \underline{37.37} \\
 & $F_{DP}\downarrow$   & 17.34 & 12.60 & 18.79 & 22.24 & 20.63 & 14.79 & 19.52 & 21.95 & 22.44 & 12.92 & 12.69 & 17.71 & 31.53 & \underline{12.18} & \textbf{11.92} \\
 & $es-AUC\uparrow$  & 36.96 & 34.48 & 37.32 & 30.05 & 36.38 & 37.35 & 32.49 & 36.91 & 37.23 & 36.42 & 36.63 & 37.42 & 31.39 & \underline{37.72} & \textbf{39.03} \\
\midrule

Overall & $AUC\uparrow$ & 56.13 & 61.35 & 60.96 & 57.39 & 60.65 & \underline{61.84} & 59.73 & 60.55 & 61.58 & 59.13 & 56.90 & 59.64 & 59.25 & 59.08 & \textbf{61.86} \\
\bottomrule
\end{tabular}%
}
\end{table*}
\label{tab:supp_dfdc}


\begin{table*}[t]
\centering
\scriptsize
\setlength{\tabcolsep}{2.2pt}
\renewcommand{\arraystretch}{1.15}
\caption{Additional experimental results on the Celeb-DF dataset with more included methods. Best results are in \textbf{bold} and second best are \underline{underlined}.}
\resizebox{\textwidth}{!}{%
\begin{tabular}{llccccccccccccccc}
\toprule
\multicolumn{2}{c}{\multirow{2}{*}{Fairness Metric}} &
\multicolumn{6}{c}{Spatial-based} &
\multicolumn{3}{c}{Frequency-based} &
\multicolumn{6}{c}{Fairness-enhanced} \\
\cmidrule(lr){3-8}\cmidrule(lr){9-11}\cmidrule(lr){12-17}
\multicolumn{2}{c}{} &
Xception & RECCE & UCF & Capsule & FFD & CORE &
F3Net & SPSL & SRM &
DAG & DAW & PFGDFD & Fairadapter & RSEF-FDD & Ours \\
\midrule

\multirow{3}{*}{Gender} & $F_{FPR}\downarrow$ & 6.93 & 13.27 & 5.70 & 20.74 & 9.19 & 5.05 & 9.24 & 15.52 & \textbf{1.18} & 8.70 & 8.71 & 7.95 & 8.59 & \underline{1.94} & 6.41 \\
 & $F_{DP}\downarrow$ & 20.23 & 14.69 & 27.82 & 12.55 & 12.84 & 15.52 & 13.62 & 22.16 & \underline{10.90} & 14.25 & 12.73 & 16.29 & 26.84 & 35.63 & \textbf{10.81} \\
 & $es-AUC\uparrow$ & 60.94 & 61.27 & 65.23 & 68.92 & 65.01 & 67.51 & 63.77 & 64.51 & 61.66 & 66.23 & 68.31 & \underline{69.47} & 56.84 & 68.92 & \textbf{69.68} \\
\midrule

\multirow{3}{*}{Race} & $F_{FPR}\downarrow$ & 23.44 & 28.41 & 36.72 & 34.69 & 17.55 & 17.78 & 16.80 & 17.49 & 31.58 & 19.44 & \textbf{14.76} & 23.99 & 46.14 & 17.40 & \underline{16.60} \\
 & $F_{DP}\downarrow$ & 17.89 & 26.26 & 21.27 & 34.26 & 22.72 & 20.97 & 19.84 & 20.01 & 10.42 & 13.26 & \underline{7.80} & 12.47 & 36.35 & 27.48 & \textbf{7.29} \\
 & $es-AUC\uparrow$ & 61.21 & 59.40 & 66.41 & 60.61 & 66.12 & \underline{69.86} & 61.35 & 63.63 & 51.73 & 62.61 & 66.47 & 67.78 & 58.34 & 67.80 & \textbf{69.96} \\
\midrule

\multirow{3}{*}{Intersection} & $F_{FPR}\downarrow$ & 27.38 & 31.69 & 51.71 & 62.46 & 30.52 & 30.04 & 27.90 & 33.32 & 23.81 & \underline{23.60} & 24.75 & 23.71 & 47.03 & 24.17 & \textbf{23.32} \\
 & $F_{DP}\downarrow$ & 18.72 & 39.04 & 26.00 & 52.04 & 38.39 & 27.40 & 29.26 & 30.17 & 16.75 & 13.80 & 13.78 & \underline{13.30} & 35.69 & 27.73 & \textbf{12.01} \\
 & $es-AUC\uparrow$ & 61.22 & 62.24 & \textbf{63.52} & 61.67 & 62.63 & 62.87 & 62.44 & 61.31 & 57.34 & 63.08 & 59.24 & 62.63 & 54.77 & 62.56 & \underline{63.27} \\
\midrule

Overall & $AUC\uparrow$ & 69.18 & 73.96 & 73.21 & 72.47 & 74.22 & 70.03 & \underline{76.70} & 75.58 & 73.21 & 72.29 & 74.09 & 74.24 & 64.94 & 74.36 & \textbf{76.75} \\
\bottomrule
\end{tabular}%
}
\end{table*}
\label{tab:supp_celeb}

\end{document}